%% file: main.tex
\documentclass[lettersize,journal]{IEEEtran}
\usepackage{amsmath,amsfonts}
\usepackage{algorithmic}
\usepackage{array}
\usepackage{textcomp}
\usepackage{stfloats}
\usepackage{url}
\usepackage{verbatim}
\usepackage{graphicx}
\usepackage{gensymb}
\usepackage{cite}
\usepackage{caption}
\usepackage{subcaption}
\def\BibTeX{{\rm B\kern-.05em{\sc i\kern-.025em b}\kern-.08em
    T\kern-.1667em\lower.7ex\hbox{E}\kern-.125emX}}
\usepackage{balance}

\newcommand{\bb}{\mathbf{b}}

\newcommand{\xb}{\mathbf{x}}

\newcommand{\ub}{\mathbf{u}}

\newcommand{\wb}{\mathbf{w}}
\newcommand{\rb}{\mathbf{r}}

\newcommand{\yb}{\mathbf{y}}

\newcommand{\ssb}{\mathbf{s}}

\DeclareMathOperator*{\argmin}{arg\,min}

\usepackage{accents}
\newcommand{\ubar}[1]{\underaccent{\bar}{#1}}

\usepackage{color}
\newcommand{\rev}[1]{{\color{red} #1}}

\begin{document}
\title{3D Path-Following Guidance via Nonlinear Model Predictive Control for FW-sUAS}

\author{C. Alexander Hirst$^{1,3}$, Chris Reale$^{2}$ and Eric Frew$^{1}$
\thanks{$^{1}$Ann and H.J. Smead Department of Aerospace Engineering Sciences at the University of Colorado Boulder, Boulder, Colorado 80303, USA
        {\tt\small camron.hirst@colorado.edu, eric.frew@colorado.edu}}%
\thanks{$^{2}$Charles Stark Draper Laboratory Inc.,
        Boston, Massachusetts 02139, USA
        {\tt\small creale@draper.com}}%
\thanks{$^{3}$ Draper Scholar}%
}


\maketitle

\begin{abstract}
\textbf{This paper presents the design, implementation, and flight test results of two novel 3D path-following guidance algorithms based on nonlinear model predictive control (MPC), with specific application to fixed-wing small uncrewed aircraft systems. To enable MPC, control-augmented modelling and system identification of the RAAVEN small uncrewed aircraft is presented. Two formulations of MPC are then showcased. The first schedules a static reference path rate over the MPC horizon, incentivizing a constant inertial speed. The second, with inspiration from model predictive contouring control, dynamically optimizes for the reference path rate over the controller horizon as the system operates. This allows for a weighted tradeoff between path progression and distance from path, two competing objectives in path-following guidance. Both controllers are formulated to operate over general smooth 3D arc-length parameterized curves. The MPC guidance algorithms are flown over several high-curvature test paths, with comparison to a baseline lookahead guidance law. The results showcase the real-world feasibility and superior performance of nonlinear MPC for 3D path-following guidance at ground speeds up to 36 $\text{m}/\text{s}$.}
\end{abstract}

\section{Introduction}
\IEEEPARstart{F}{ixed-wing} small uncrewed aircraft systems (FW-sUAS) are aerial robotic platforms which have broad use cases in a variety of academic and commercial applications. FW-sUAS generally exhibit greater payload capacity, endurance, airspeed, and robustness compared to multirotors, desirable characteristics for many types of field missions. However, due to nonlinear dynamics, high inertia, low thrust-to-weight ratios, imprecise sensors and actuators, wind disturbances, and onboard computational constraints, high-performance path-following guidance for FW-sUAS is challenging. It follows that common FW-sUAS guidance algorithms are generally closed-form expressions designed for simple classes of paths (i.e. Dubin's paths \cite{beard2012small}), and exhibit conservative performance. Furthermore, airspeed setpoints are assumed stationary and prescribed by a human operator, limiting performance in terms of rapid path progression.

While modern robotic planning algorithms can generate high-utility 3D path solutions for complex missions \cite{mills2017energy}, execution of such paths remains a challenge with current guidance methods. This paper addresses a fundamental challenge in the control stack for autonomous FW-sUAS: given a smooth 3D curve (i.e. path) to follow, formulate a guidance algorithm which minimizes distance from the path while maximizing path progression. This objective has significant relevance in time-critical applications such as search-and-rescue and atmospheric science.

\begin{figure}
    \centering
    \includegraphics[width=\columnwidth]{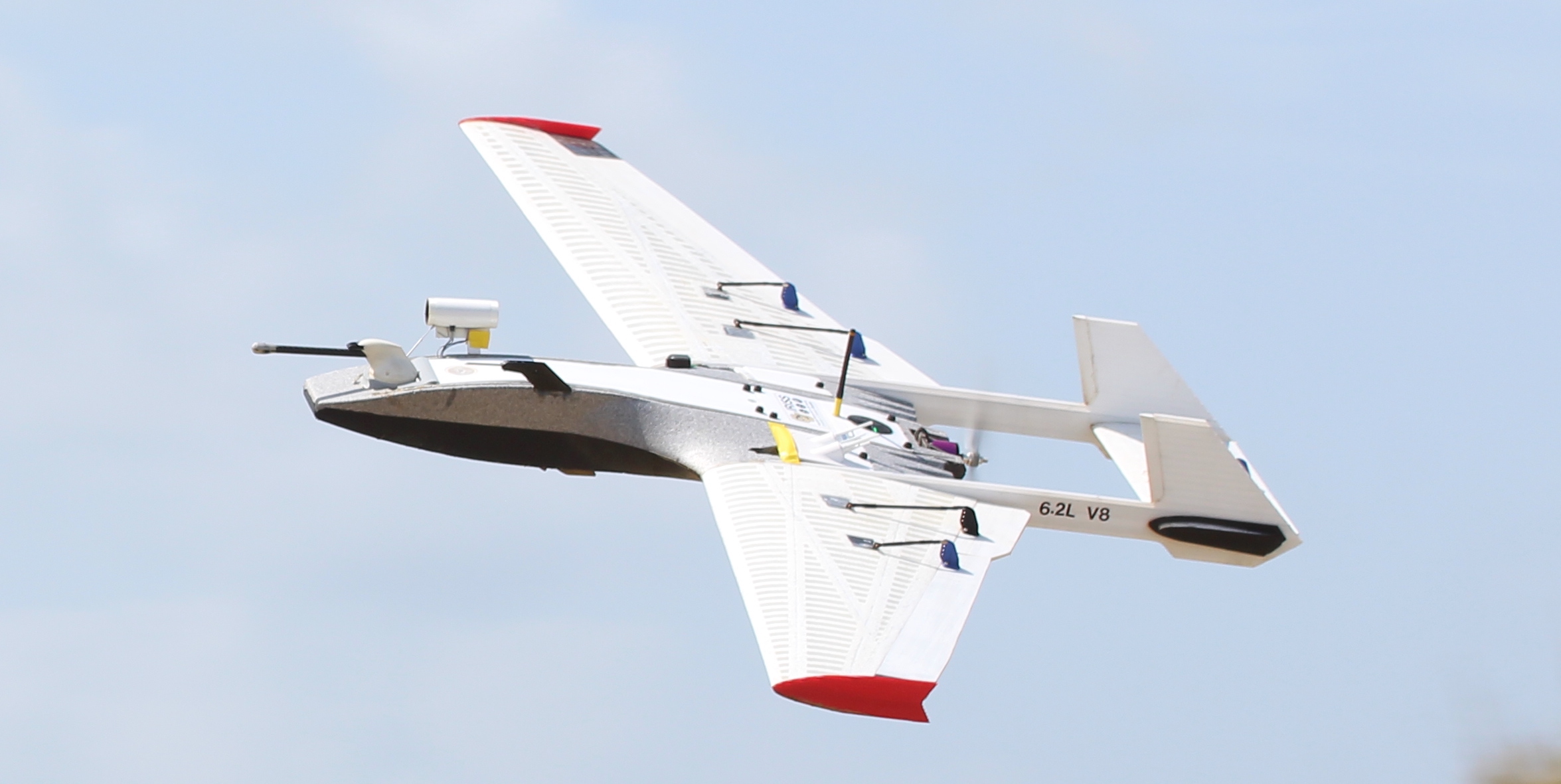}
    \caption{A RAAVEN aircraft, outfitted with an array of atmospheric sensors, conducting a low-pass maneuver during a scientific field campaign.}
    \label{fig:raaven}
\end{figure}

To achieve fast and accurate path following, this work leverages nonlinear model predictive control (MPC) to generate near-optimal guidance commands in real time. Nonlinear MPC enables simultaneous handling of nonlinear dynamics, state/input constraints, and non-myopic control horizons, all of which are critical to enable safe, efficient, and complex behaviors. MPC has risen to prominence as a state-of-the-art methodology for various robotic tasks, such as legged locomotion \cite{neunert2018whole}, agile quadrotor flight \cite{romero2022model}, and autonomous car racing\cite{liniger2015optimization}. This work builds upon prior work on MPC for FW-sUAS guidance \cite{stastny2018nonlinear,hirst2022nonlinear} to enable high-performance 3D path-following guidance over generic curves. As a preliminary step, control-augmented modelling and system identification results for the RAAVEN FW-sUAS (Figure \ref{fig:raaven}) are presented. Then, two novel formulations of path-following MPC for FW-sUAS are presented in detail. The first prescribes a constant reference path rate to define the MPC cost function. The second, inspired by prior works on model predictive contouring control \cite{lam2010model,brito2019model,romero2022model}, incorporates the reference path rate as a decision variable within MPC. Finally, we deploy both MPC algorithms onboard an embedded companion computer in real-world flight tests to rigorously evaluate the feasibility and utility on a variety of 3D paths. Comparisons to a baseline lookahead controller \cite{bird2014closing,park2007performance} reveal the superior performance of MPC for FW-sUAS guidance. In summary, this paper makes the following contributions:

\begin{enumerate}
    \item A control-augmented model of the RAAVEN FW-sUAS.
    \item A detailed formulation of two novel 3D path-following MPC guidance algorithms for use with FW-sUAS along with descriptions of their implementations.
    \item Fielded experiments to demonstrate the real-world feasibility and superior performance of these guidance algorithms on a set of challenging paths.
\end{enumerate}

\subsection{Related Work}

Literature on path-following methods for FW-sUAS is vast and includes geometric, vector-field, and LQR-based approaches. For more information, we refer the interested reader to surveys on 2D and 3D path-following guidance methods for FW-sUAS  \cite{sujit2014unmanned,pelizer2017comparison}. These approaches, while robust and computationally efficient, are conservative to ensure the vehicle remains in a safe flight envelope. As a result, these methods generally do not explicitly reason about the aircraft dynamics or limitations.

For common robotic applications such as industrial arms, quadrotors, and ground vehicles, nonlinear MPC has achieved significant attention as a high-performance path-following strategy \cite{faulwasser2015nonlinear,diwale2017model,faulwasser2016implementation,gros2012aircraft,xu2019design, kabzan2019learning, romero2022model}. Theoretical requirements for stability and recursive feasibility were developed via terminal costs and constraints \cite{faulwasser2015nonlinear} and further refined for an airborne wind energy system \cite{diwale2017model}. The authors then used a \textit{truncated} path-following MPC scheme (no provably stabilizing terminal costs or constraints) to demonstrate the approach onboard a robotic arm \cite{faulwasser2016implementation}. Stability was empirically demonstrated by choosing a sufficiently long MPC horizon \cite{gros2012aircraft, grune2017nonlinear}. Due to the difficulty in deriving stabilizing terms, truncated path-following MPC formulations are often applied for real-world applications onboard complex vehicles, exhibiting stability and recursive feasibility in practice \cite{liniger2015optimization, kayacan2015robust, brito2019model, xu2019design, kabzan2019learning, romero2022model}. This work pairs the truncated MPC approach with long horizons to demonstrate high-performance path-following MPC onboard FW-sUAS.

Nonlinear model predictive control for FW-sUAS guidance has received limited attention from the research community over the last two decades. Early work on nonlinear MPC for FW-sUAS largely focused on 2D lateral control \cite{kang2009linear, yang2013adaptive} with simple kinematic models, tested in simulated environments. 

Later works extended this idea to three-dimensional guidance, with an emphasis on increasing closed-loop performance and exploiting more of the flight envelope. It has been demonstrated that MPC could conduct fast turn and flip maneuvers, as well as obstacle avoidance \cite{gros2012aircraft} . Lookahead guidance laws have been used to hot-start an iterative optimization algorithm \cite{gavilan2015iterative}. That approach enabled fast convergence of the algorithm, while maintaining a feasible backup solution in case of non-convergence. Sampling-based optimization was proposed to enable aggressive flight through a series of waypoints \cite{pravitra2021flying}. While effective, the algorithm was not demonstrated to be real-time feasible on an embedded computer. Recently, a different approach was taken, where MPC was used to perform path-following guidance through direct actuator control \cite{reinhardt2023fixed}. While that approach allows for lower-level integration, low-level predictive control requires detailed aerodynamic modelling which is time consuming and expensive. Furthermore, actuator-level update rates are commonly on the order of 100 Hz for FW-sUAS, a challenge for MPC on embedded hardware \cite{bohn2019deep}. A shortcoming of those prior 3D MPC guidance works \cite{gros2012aircraft, gavilan2015iterative, pravitra2021flying, reinhardt2023fixed} is the lack of flight testing with real-world disturbances, computation constraints, sensor noise, and plant-model mismatch. 

The singular work illustrating 3D path-following MPC guidance for FW-sUAS in windy environments via real-world flight experiments is \cite{stastny2018nonlinear}. The authors developed a control-augmented guidance model, which included inner-loop attitude stabilization. By controlling the aircraft at a higher level, guidance commands are updated at slower rates (10-20 Hz), freeing up precious time and computational resources. However, the MPC formulation relies on a lookahead guidance algorithm \cite{cho2015three} to generate course and flight path angle references within MPC. That approach implicitly confines the performance of the algorithm, while also limiting the applicability to paths composed of lines, circles, and helices. Furthermore, airspeed is regulated to a constant reference, which simplifies the optimization problem but also limits performance.

With an eye towards enabling fast, precise path-following guidance, this paper directly builds off \cite{stastny2018nonlinear} to develop, implement and assess two novel 3D path-following MPC guidance algorithms for FW-sUAS in windy environments. The algorithms are formulated to work with arbitrary continuous arc-length parameterized paths defined in position space. This relaxation greatly increases the utility of the guidance algorithm in demanding applications. The first algorithm, denoted constant rate MPC (CR-MPC) attempts to stay on the path while the reference path parameter moves at a pre-defined, constant inertial speed. The second algorithm takes inspiration from model predictive contouring control (MPCC) literature \cite{liniger2015optimization,romero2022model}, and includes the reference path parameter rate as a decision variable within the optimization problem itself. This inclusion enables MPCC to directly tradeoff maintaining high airspeeds while minimizing path error. Weighted tradeoff between objectives such as this cannot be achieved with standard guidance algorithms.

To enable real-world flight tests of CR-MPC and MPCC, a new control-augmented model of the RAAVEN FW-sUAS \cite{frew2020field} is formulated and fitted to flight test data. Finally, CR-MPC and MPCC are flown over four paths of varying difficulty, and compared to the lookahead guidance algorithm presented in \cite{bird2014closing}.


\section{Aircraft-Autopilot Model}

This work utilizes a control-augmented dynamics model to capture aircraft dynamics when combined with a low-level autopilot conducting attitude stabilization \cite{stastny2018nonlinear}. Besides allowing for lower update rates, the control-augmented aircraft model is admissible to system identification via relatively small amounts of flight test data. This presents a significant advantage for practitioners without access to large wind tunnel facilities for system identification, as in \cite{gryte2018aerodynamic}. 

\begin{figure}
    \centering
    \includegraphics[width=\columnwidth]{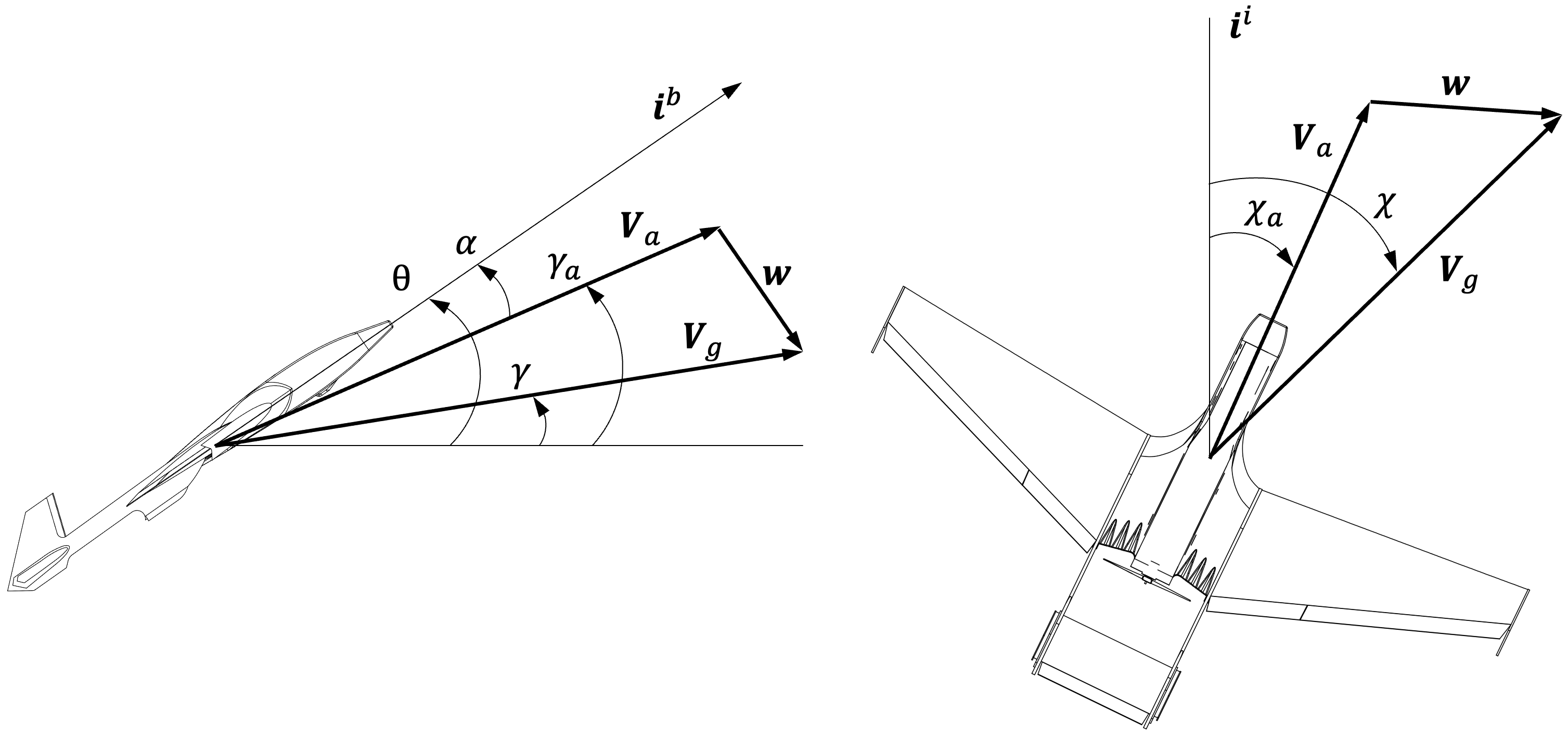}
    \caption{Angle definitions with respect to the aircraft body \textbf{\textit{b}} and inertial \textbf{\textit{i}} frame assuming side-slip angle $\beta = 0$. $\mathbf{V}_a$ and $\mathbf{V}_g$ are the air-relative and ground velocity vectors, respectively.}
    \label{fig:raaven_angles}
\end{figure}

The aircraft utilized in this work is the RAAVEN FW-sUAS (Figure \ref{fig:raaven}), a pusher propeller, twin-boom aircraft developed for in-situ atmospheric sensing of severe weather. The RAAVEN has been deployed in numerous scientific field campaigns \cite{frew2020field}. The aircraft weighs about 6.65 kg (depending on payload), with a maximum level airspeed of 40 m/s. Maximum endurance is 3 hours in ideal conditions utilizing a 621 Wh battery. The aircraft is equipped with a Cube Orange flight computer running the open-source PX4 v1.13.1 autopilot firmware. The PX4 autopilot provides high-rate aircraft state estimates, 2D wind estimates, and low-level attitude stabilization. Utilizing PX4 distinguishes this work from prior work \cite{stastny2018nonlinear}, which uses custom EKF and attitude controllers.


Nominally, PX4 performs path-following guidance by decoupling the lateral and longitudinal control objectives. Lookahead guidance \cite{park2007performance} is used to command lateral accelerations, which are converted to roll angle setpoints for the low-level attitude controller. Altitude and airspeed are controlled via the total energy control system (TECS) algorithm \cite{bruce1986nasa} via pitch angle and throttle setpoints. The MPC guidance controllers in this work replace both TECS and lookahead guidance by reasoning over the aircraft attitude-controlled dynamics, longitudinal dynamics and coupled lateral-longitudinal kinematics.

The control-augmented aircraft model takes the form:

\begin{equation}\label{eq:aa_model}
    \begin{bmatrix}
        \Dot{n} \\
        \Dot{e} \\
        \Dot{d} \\
        \Dot{\phi} \\
        \Dot{\theta} \\
        \Dot{\chi_a} \\
        \Dot{V_a} \\
        \Dot{\gamma_a} \\
        \Dot{\delta_T}
    \end{bmatrix} =
    \begin{bmatrix}
        V_a \cos{\gamma_a} \cos{\chi_a} + w_n \\  
        V_a \cos{\gamma_a} \sin{\chi_a} + w_e \\  
        -V_a \sin{\gamma_a} + w_d \\  
        K_\phi (\phi_c - \phi) \\ 
        K_\theta (\theta_c  - \theta) \\ 
        \sin{\phi} ( T \sin{\alpha} + L) / (m V_a \cos{\gamma_a}) \\  
        (T \cos{\alpha} - D)/m - g \sin{\gamma_a} \\ 
        [(T \sin{\alpha} + L)\cos{\phi} - mg\cos{\gamma_a}] / (m V_a) \\  
        (\delta_{T,c} - \delta_T) / \tau_T
    \end{bmatrix}
\end{equation}

\noindent where $n, e, d$ are inertial 3D aircraft position components in the north-east-down frame. Wind vector $\mathbf{w} = [w_n, w_e, w_d] ^ \top$ is composed of north, east, and down components. $V_a$ and $\gamma_a$ are the (true) airspeed and air-relative flight path angles, respectively. Roll and pitch angles $\phi, \theta$ are regulated by the low-level attitude controller to commanded values $\phi_c, \theta_c$. FW-sUAS autopilots such as PX4 commonly attempt to achieve first order response in attitude setpoints \cite{bird2014closing}, and are modelled accordingly. As in \cite{stastny2018nonlinear}, it is also assumed that the autopilot effectively regulates sideslip to $\beta = 0$, which means air-relative heading $\chi_a$ is approximately equal to the aircraft heading angle (Figure \ref{fig:raaven_angles}). Furthermore this implies that the angle of attack $\alpha$ is related to  $\gamma_a$ and $\theta$ by: $\alpha \approx \theta - \gamma_a$. Note that $\gamma_a$ is not directly measured, but calculated via $\gamma_a = \arcsin(-1(\Dot{d} - w_d)/V_a)$ \cite{beard2012small}. This work assumes that the underlying autopilot for the RAAVEN under nominal flight conditions maintains $V_a > 0$ and $|\gamma_a| < \pi / 2$, ensuring that (\ref{eq:aa_model}) is well defined. See Section \ref{section:cr-mpc} for additional details.

Aircraft mass $m$ and gravitational constant $g$ are assumed known and constant. The aircraft virtual throttle state $\delta_{T}$ is also assumed driven to commanded throttle state $\delta_{T,c}$ via first-order dynamics. Therefore, the control vector is defined as $\ub = [\phi_c, \theta_c, \delta_{T,c}]^\top$. Roll and pitch constants $K_\phi$ and $ K_\theta$ are the closed-loop parameters to be fit to the aircraft-autopilot response. These closed-loop parameters should be refit if any changes in the underlying attitude controller occur. 

The aircraft lift $L$, drag $D$, and thrust $T$ forces \cite{beard2012small,coates2019propulsion} are modelled as:

\begin{equation}\label{eq:LDT_forces}
    \begin{bmatrix}
        L \\
        D \\
        T
    \end{bmatrix} =
    \begin{bmatrix}
        \frac{1}{2} \rho V_a^2 S (C_{L0} + C_{L1} \alpha) \\  
        \frac{1}{2} \rho V_a^2 S (C_{D0} + C_{D1} \alpha + C_{D2} \alpha^2) \\  
        \rho S_{p} C_{T} \delta_T ( V_{\infty} + \delta_T (k_m - V_{\infty}))(k_m - V_{\infty})  
    \end{bmatrix}
\end{equation}

\noindent where $\rho, S, S_p$ are air density, wing area, and propeller swept area, respectively. For the RAAVEN, $S$ and $S_p$ are 1.02$\text{ m}^2$ and 0.0856$\text{ m}^2$, respectively. The free-stream velocity is defined as $V_{\infty} = V_a \cos{\alpha}$. The ``open-loop" lift parameters $C_{L0}, C_{L1}$, drag parameters $C_{D0}, C_{D1}, C_{D2}$, thrust constant $C_{T}$, throttle time constant $\tau_T$, and motor constant $k_m$ are all simultaneously regressed from flight data. These parameters, once fit to an aircraft, should be independent of changes in the attitude controller. As direct measurement of lift, drag, and thrust forces is infeasible directly from flight data, these variables are related to the body-frame acceleration observations (readily available from the onboard IMU) via \cite{beard2012small}:  

\begin{equation}\label{eq:imu_eqns}
    \begin{bmatrix}
        a_x \\
        a_z
    \end{bmatrix} =
    \begin{bmatrix}
        \cos{\alpha}  & \sin{\alpha} \\ 
        \sin{\alpha} & -\cos{\alpha}  
    \end{bmatrix}
        \begin{bmatrix}
        (T \cos{\alpha} - D ) / m \\ 
        (T \sin{\alpha} + L ) / m 
    \end{bmatrix}.
\end{equation}

In summary, the control-augmented aircraft model (\ref{eq:aa_model}) used for guidance-level control requires fitting closed-loop parameters $\mathbf{p}_c = [K_\phi, K_\theta]^\top$ and open-loop parameters $\mathbf{p}_o = [\tau_T, C_{L0}, C_{L1}, C_{D0}, C_{D1}, C_{D2}, C_{T}, k_m]^\top$.

These parameters were fit to data collected during a series of test flights. Data was collected via the onboard companion computer at 40 Hz. In total, 38 minutes of flight data was collected for system identification. Data was separated into an 80/20 training/testing split. Training data consisted of piloted static and 2-1-1 maneuvers in the PX4 stabilized flight mode \cite{stastny2018nonlinear}. While these system identification maneuvers can be automated, the limited area available for flight testing and relatively high airspeeds of the RAAVEN made safe execution of open-loop system identification maneuvers cumbersome. Flight data consisted of roll angles between $[-45\degree,45\degree]$, pitch angles between $[-20\degree, 20\degree]$, and airspeeds between $[15, 40]$ m/s. The validation dataset consisted of ``free-form" maneuvers generated by the pilot attempting to explore the complete flight envelope of the aircraft. PX4 EKF estimates were logged directly and used for system identification without any additional processing.

\begin{figure}
    \centering
    \includegraphics[width=\columnwidth]{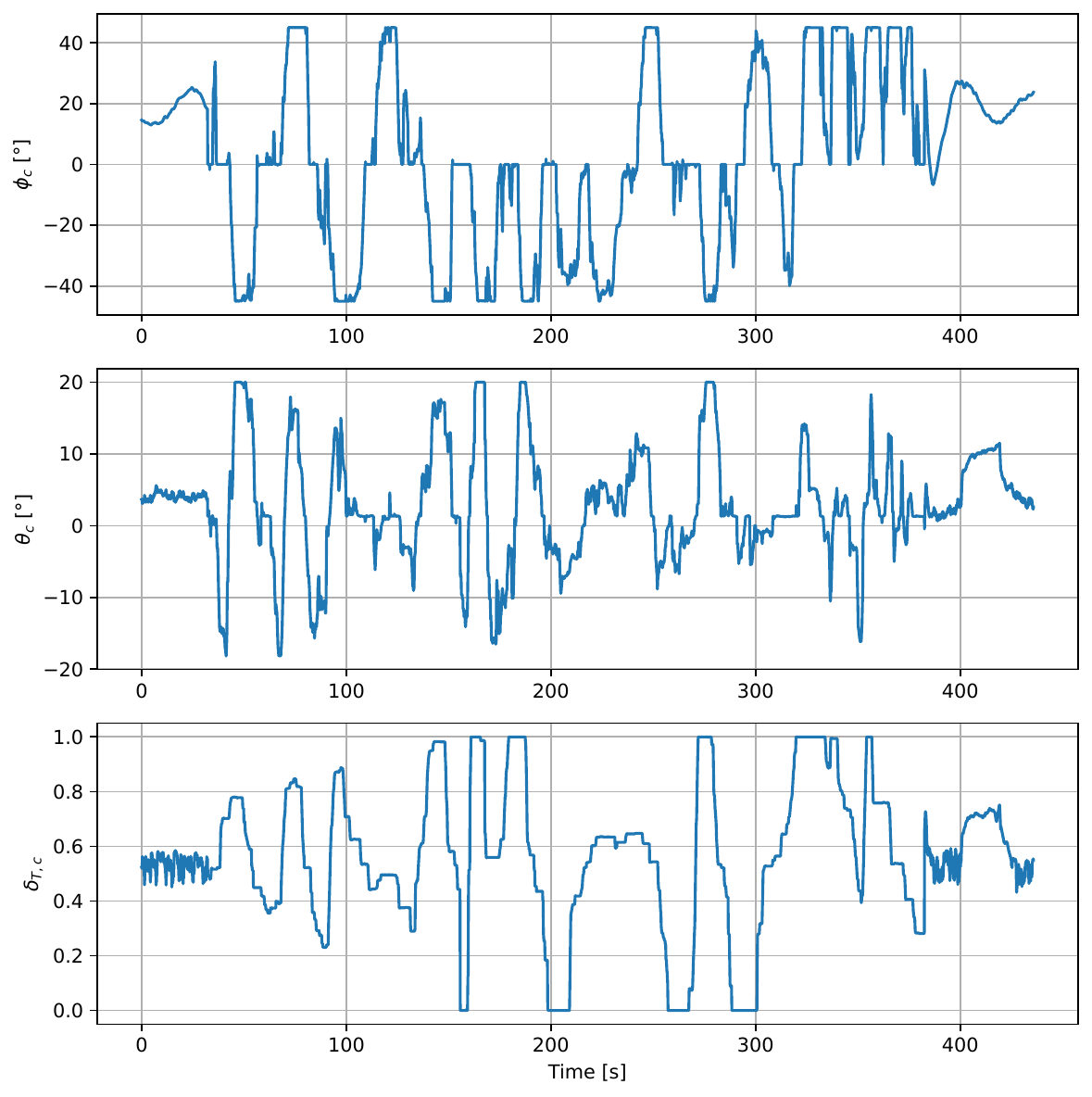}
    \caption{Control inputs during free-form validation flight.}
    \label{fig:m4_validation_controls}
\end{figure}

\begin{figure}
    \centering
    \includegraphics[width=\columnwidth]{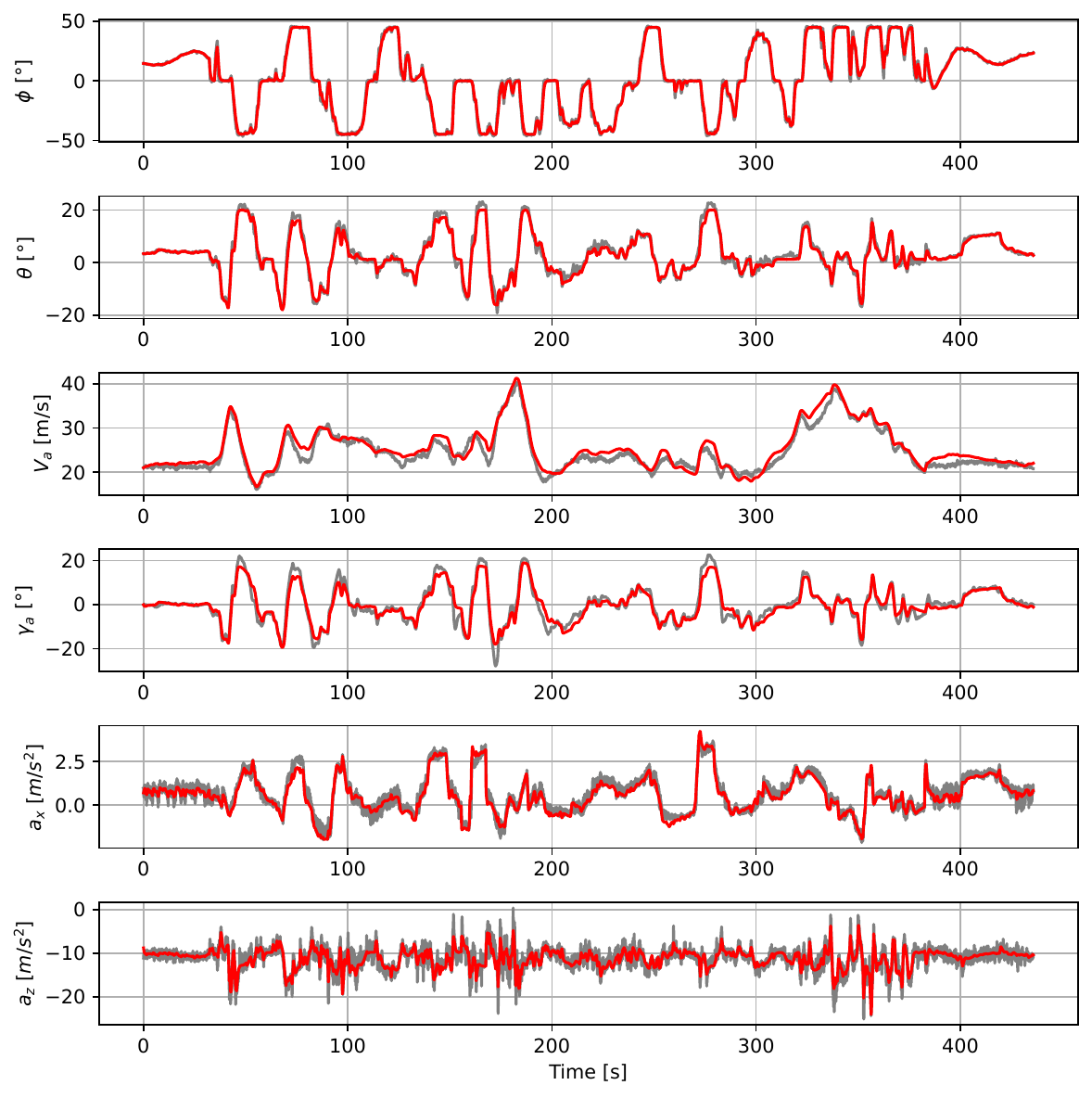}
    \caption{Control-augmented aircraft model output (red) on validation flight data (grey).}
    \label{fig:m4_validation_fit}
\end{figure}

To fit the parameters to the training data, the MATLAB system identification toolbox was used. A grey-box model was formulated for each of the decoupled open-loop and closed-loop parameters, as in \cite{stastny2018nonlinear}. The parameters were fitted using the default optimization algorithm \texttt{lsqnonlin}, which is a trust-region-reflective algorithm for nonlinear least-squares optimization problems. After open-loop and closed-loop models were fit, they were combined into the complete control-augmented aircraft model (\ref{eq:aa_model}) for validation testing on the free-form dataset (Figures \ref{fig:m4_validation_controls}, \ref{fig:m4_validation_fit}). RMSE values for each output variable are shown in Table \ref{tab:validation_rmse}. Table \ref{tab:fitted_params} contains the fitted model parameters. It is noted that while parameters fitted to aerodynamic equations (\ref{eq:LDT_forces}) may admit a low-cost fit via output-error grey-box modelling, the resulting parameters may not represent physical quantities and should be interpreted appropriately.

\begin{table}[]
\centering
\caption{RMSE values of the control-augmented model over the free-form validation dataset.}
\label{tab:validation_rmse}
\resizebox{0.9\columnwidth}{!}{%
\begin{tabular}{ll|ll}
Variable              & RMSE       & Variable                 & RMSE                           \\ \hline
$\phi$   & 2.504\degree  & $\gamma_a$ & 2.3296\degree                     \\
$\theta$ & 1.4359\degree & $a_x$      & 0.2736 $\text{m}/\text{s}^2$ \\
$V_a$    & 1.3744 m/s & $a_z$      & 1.2113 $\text{m}/\text{s}^2$ 
\end{tabular}%
}
\end{table}



\begin{table}[]
\centering
\caption{Fitted model parameters of the RAAVEN FW-sUAS control-augmented system from the output-error system identification procedure.}
\label{tab:fitted_params}
\resizebox{0.85\columnwidth}{!}{%
\begin{tabular}{ll|ll}
Parameter & Value    & Parameter & Value  \\ \hline
$\tau_T$    & 0.1161   & $C_{L0}$     & 0.0917 \\
$C_T$      & 0.0233    & $C_{L1}$     & 2.7493 \\
$k_m$      & 143.3052 & $K_\phi$   & 2.0316 \\
$C_{D0}$   & 0.0362   & $K_\theta$  & 2.1498 \\
$C_{D1}$   & 0.0868   & $C_{D2}$  & 0.4459 
\end{tabular}%
}
\end{table}

\section{Nonlinear MPC for Path-Following Guidance}

The objective of this work is to develop and test model predictive control algorithms for high-performance 3D path-following guidance onboard FW-sUAS.

To begin, consider a known reference path with path parameter $\psi \in  \Psi \subset \mathbb{R}$. Define a path as the mapping $\rb_P : \Psi \rightarrow R$ from path parameter space to 3D position space $R \subset \mathbb{R}^3$. As in previous works, paths are assumed to be pre-processed into arc-length parameterized $\mathcal{C}^2$ continuous B-splines \cite{romero2022model,hirst2022nonlinear}. Given the aircraft inertial position $\rb_i = [n, e, d]^\top$ and reference path parameter $\psi_i$ at time $t_i$, the path error vector is $[\epsilon_n, \epsilon_e, \epsilon_d]^\top = \rb_i - \rb_P(\psi_i)$. Often, $\psi_i$ is chosen to be the closest point on the path to the vehicle at time $t_i$. Define the closest path parameter to be:

\begin{equation} \label{optimal_psi}
    \psi^*(\rb_i) = \psi^*_i = \argmin_{\psi \in \Psi} ||\rb_i - \rb_P(\psi)||_2 .
\end{equation}

Analytical evaluation of (\ref{optimal_psi}) is feasible for simple paths, such as circles or lines. However for generic paths, (\ref{optimal_psi}) is a non-convex optimization problem, and thus cannot be efficiently calculated within MPC over the entire OCP horizon during the optimization procedure. Therefore, denote a \textit{chosen} reference path parameter as $\Hat{\psi}$, which must be specified for each stage in the optimal control problem horizon. 

Position tracking assumes a known path reference $\hat{\psi}$ over all timesteps at which the system is tasked with tracking. For arc-length parameterized paths, this translates to a known inertial speed of the path reference over time. While position tracking can be effective for robotic systems experiencing small disturbances, FW-sUAS often operate in strong and dynamic wind conditions, making a priori scheduling of $\hat{\psi}$ at each timestep over the entire reference path challenging. As an alternative to position tracking, path following does not enforce \textit{when to be where} on $\rb_P$ \cite{faulwasser2016implementation}. This work develops two path-following MPC algorithms: CR-MPC and MPCC.

Inspired by position tracking, we denote the first class of path-following MPC algorithms as constant (path parameter) rate model predictive control (CR-MPC). CR-MPC assumes a known path reference rate $\Dot{\hat{\psi}}$ at which the system is tasked with following over the control horizon. CR-MPC differs from position tracking by calculating the closest path parameter for use as the initial reference path parameter before each query. This allows CR-MPC to continuously adapt when the schedule is not feasible, but still incentivize a constant inertial speed at which to follow the path.

A second method for predictive path following is model predictive contouring control (MPCC) \cite{lam2010model,brito2019model,romero2022model}. MPCC methods assign $\hat{\psi}$ and its derivative(s) as decision variables within the optimal control problem, providing online flexibility at the cost of additional computational resources. MPCC is useful for applications in which large changes of the path reference rate (and therefore inertial speed) are desired. An adaptive reference path rate also allows MPCC to explicitly trade off between speed and path-following error within the OCP objective.

We now provide some preliminary definitions before defining the CR-MPC and MPCC optimal control problems (OCPs) for FW-sUAS path following. Define the control-augmented aircraft state vector as $\xb \in \mathcal{X} \subset \mathbb{R}^9$ and control vector as $\ub \in \mathcal{U} \subset \mathbb{R}^3$. The continuous-time dynamics (\ref{eq:aa_model}) are discretized via RK4 integration into the discrete-time mapping $\xb_{k+1} = F_x(\xb_k, \ub_k, \wb_k)$ where $k \in \mathbb{Z}$ denotes the discrete time index and $\Delta t \in \mathbb{R}$ is the discrete timestep. Utilizing the direct multiple-shooting scheme \cite{rawlings2017model}, the generic finite-horizon optimal control problem is written as:

\begin{align}
\begin{split}
    &\argmin_{\mathbf{X}, \mathbf{U}} \sum_{k = 0}^{N-1} J_k(\xb_{k+1},\ub_k)\\
    & \text{subject to:} \\
    & \xb_{k+1} = F_x(\xb_k, \ub_k, \wb_k)\\
    & \xb_{k=0}  = \xb_i, \quad \wb_k = \wb_i \\
    & h_k(\xb_{k+1},  \ub_{k})\leq 0 \\
    & \forall k \in (0,\dots,N-1) \\    
\end{split}
\label{direct_multi_shoot}
\end{align}

\noindent where $J_k \, \forall \, k \in [0, \dots, N-1]$ are stage cost functions and $\xb_i, \wb_i$ are the state and disturbance realization at initial time $t_i$. The state and control trajectories are denoted as $\mathbf{X} \in \mathcal{X}^{N+1}$ and $\mathbf{U} \in \mathcal{U}^N$. Equality/inequality constraints at each stage are written as $h_k \, \forall \, k \in [0, \dots, N-1]$. Given the highly nonlinear aircraft dynamics $F_x$, it is immediately observed that (\ref{direct_multi_shoot}) is a nonconvex optimization algorithm, therefore locally optimal solutions can be found using standard optimization routines. Solving (\ref{direct_multi_shoot}) returns a (locally) optimal sequence of open-loop control actions $(\ub_0^*, \dots, \ub_{N-1}^*) = \mathbf{U}^*_{i} \in \mathcal{U}^N$, of which the first action $\ub_0^*$ is applied to the plant in a receding-horizon fashion in MPC frameworks. Therefore, the MPC control policy is defined as:

\begin{equation}
     \pi_{MPC}(\xb_i, \wb_i) = \ub^*_{0}. 
\end{equation}

\rev{} The next sections detail the implementation of the CR-MPC and MPCC optimal control problems. 

\subsection{CR-MPC Optimal Control Problem}
\label{section:cr-mpc}

CR-MPC assumes a desired $\Dot{\Hat{\psi}}$ is known before querying the controller, and is fixed over the controller horizon. By calculating $\psi^*(\rb_i)$, a sequence of reference path parameters can be defined over the OCP horizon. The reference path parameter at stage $k$ is defined as:

\begin{equation}
    \hat{\psi}_k = \psi^*(\rb_i) + \Dot{\Hat{\psi}} k \Delta t   \quad \forall k \in (0, \dots, N)
\end{equation}

Given a reference path parameter $\Hat{\psi}$ at each stage and dropping the subscript $k$ for brevity, the position errors $\epsilon_n, \epsilon_e, \epsilon_d$ are defined as:

\begin{equation} \label{eq:ocp_path_error}
    [\epsilon_n, \epsilon_e, \epsilon_d]^\top
    =
    \rb - \rb_P(\Hat{\psi}).
\end{equation}

An error term on the aircraft course angle is used to align horizontal aircraft velocity with the path tangent. Given the unit tangent to the path $\mathbf{T}^P = [T^P_{n}, T^P_{e}, T^P_{d}]^\top$ at $\hat{\psi}$, the course angle error is:

\begin{equation}
    \epsilon_{C} = \arctan{(\Dot{e}, \Dot{n})} - \arctan{(T^P_{e}, T^P_{n})}
\end{equation}

\noindent where $\arctan$ is the four quadrant arc-tangent operator. $T^P_{n}$, $T^P_{e}$, $T^P_{d}$ are the unit tangent components of the path in the north, east, and down directions, respectively. The course angle error is then wrapped between $\pm \pi$ to get:

\begin{equation} \label{eq:course_angle_error}
    \epsilon_{\chi} = \arctan{(\sin{(\epsilon_{C})}, \cos{(\epsilon_{C})})}.
\end{equation}

\noindent Next, $\epsilon_{\gamma}$ is the deviation from the nominal flight path angle:

\begin{equation}
    \epsilon_{\gamma} = \gamma - \arcsin{(-T^P_{d}/ ||\mathbf{T}^P||_2)}.
\end{equation}

Slack variables $\Bar{s}_{\alpha}, \Bar{s}_V, \ubar{s}_\alpha, \ubar{s}_V$ are introduced to formulate soft constraints on the aircraft's angle of attack and airspeed. This strongly incentivizes keeping the aircraft in an admissible flight regime (i.e. preventing stall) while allowing for momentary constraint violation due to plant-model mismatch, sensor noise, or changing wind conditions \cite{stastny2018nonlinear}. The soft constraints are written as:

\begin{equation} \label{eq:soft_constr}
    \begin{split}
        V_a - \Bar{s}_V - \Bar{V}_a &\leq 0, \quad  \Bar{s}_V \geq 0 \\
        \alpha - \Bar{s}_{\alpha} -  \Bar{\alpha} & \leq 0, \quad \Bar{s}_{\alpha} \geq 0 \\
        \ubar{V}_a -  V_a - \ubar{s}_V & \leq 0, \quad \ubar{s}_V \geq 0\\
        \ubar{\alpha} - \alpha - \ubar{s}_{\alpha} & \leq 0, \quad \ubar{s}_\alpha \geq 0
    \end{split}
\end{equation}

\noindent where $[\ubar{\alpha}, \Bar{\alpha}]$ and $[\ubar{V}_a, \Bar{V}_a]$ define the admissible angle of attack and airspeed ranges, respectively. The inputs, $\ub_k = [\phi_c, \theta_c, \delta_{T,c}]$, are hard box-constrained to ensure feasible attitude/throttle commands to the autopilot. See Table \ref{tab:mpc_params} for input constraint parameters.

Finally, the magnitudes of the predicted roll, pitch and throttle rates at each intermediate stage are penalized to incentivize control inputs which are close to the current roll pitch and throttle states. These penalties helped reduce control oscillations induced by unmodelled transient dynamics of the aircraft-autopilot system during flight testing.

Given these cost components, the optimal control problem at time $t_i$ used by CR-MPC is defined in a generic nonlinear least squares form:

\begin{align}\label{eq:mpc_ocp}
\begin{split}
    \argmin_{\mathbf{X}, \mathbf{U}, \mathbf{S}} & \frac{1}{2}  \sum_{k = 1}^{N}  \begin{bmatrix} \yb_k \\ \ssb_k \end{bmatrix}^\top \begin{bmatrix} Q & 0  \\  0 & S \end{bmatrix} \begin{bmatrix} \yb_k \\  \ssb_k \end{bmatrix} \\
    &+ \frac{1}{2} \sum_{k = 0}^{N-1} \begin{bmatrix} \bb_k \\ \Delta \ub_k \end{bmatrix}^\top \begin{bmatrix} B & 0 \\ 0 & R_k \end{bmatrix} \begin{bmatrix} \bb_k \\ \Delta \ub_k \end{bmatrix} \\    & \text{subject to:} \\
    & \xb_{k+1} = F_x(\xb_k, \ub_k, \wb_k)\\
    & \xb_{k=0}  = \xb_i, \quad \wb_k = \wb_i, \quad \ub_k \in \mathcal{U} \\
    & h_k(\xb_{k},  \ub_{k}, \ssb_k) \leq 0, \quad h_N(\xb_{N}, \ssb_N)\leq 0\\
    &\forall k \in (0,\dots,N-1) \\    
\end{split}
\end{align}


\noindent where the rate vector is $\bb_k(\xb_k, \ub_k) = [\Dot{\phi}, \Dot{\theta}, \Dot{\delta_T}]^\top$, the state error vector is $\yb_k(\xb_k) = [\epsilon_n, \epsilon_e, \epsilon_d, \epsilon_\chi, \epsilon_\gamma]^\top$,  the control slew vector is $\Delta \ub_k = \ub_k - \ub^*_{k, i-1}$, and the slack variable vector is $\ssb_k = [\Bar{s}_V, \Bar{s}_\alpha, \ubar{s}_V, \ubar{s}_\alpha] ^ \top \in \mathcal{S}$. The slack variable trajectory is written as $\mathbf{S} \in \mathcal{S}^{N+1}$. $Q$, $B$, $S$ and $R_k$ are non-negative diagonal weighting matrices for the state error, rates, slack, and slew rate terms at each stage, respectively. The slew rate penalty is included to avoid bang-bang control between subsequent iterations of the MPC controller. Take $R_k = \lambda ^k R$ such that $0 \leq \lambda \leq 1$ to discount slew rate penalties further into the horizon. Discounting the slew rate penalty reduces unnecessarily sluggish performance \cite{stastny2017nonlinear}. All other weighting matricies are constant across the horizon. The soft constraint equations (\ref{eq:soft_constr}) are contained in the generic stage inequality constraints $h_k$.

This NMPC formulation does not have provable stability or recursive feasibility due to the lack of terminal cost or constraints. As noted above, this formulation is an example of a truncated MPC with a long horizon \cite{liniger2015optimization, kayacan2015robust, brito2019model, xu2019design, kabzan2019learning, romero2022model,grune2017nonlinear}. Such approaches have been empirically shown to be stable and to perform well. For the problem formulations considered here, the problem structure is such that infeasibility would only occur if the initial conditions lead to undefined dynamics in Equation (\ref{eq:aa_model}), i.e. if the airspeed $V_a = 0$ or the air-relative flight path angle $\gamma_a = \pi/2$. This work assumes airspeed soft constraints within the MPC algorithm paired with a well-tuned autopilot can maintain airspeed above the stall speed of the aircraft and hence well above zero. Assumptions could be put on the wind speed or wind gusts (change in wind speeds) to ensure that the airspeed cannot drop to zero. Such assumptions are typically part of the allowable flight envelope of any FW-sUAS. In contrast, it is not possible to constrain $\gamma_a$ through limits on wind speed or wind gusts alone. The RAAVEN is a conventional fixed-wing aircraft with low thrust-to-weight ratio and a hard constrained commanded pitch angle command $\vert \theta_c \vert \leq 10^{\circ}$. The angle of attack $\alpha$ is soft constrained to the interval $\alpha \in [-6^{\circ}, 12^{\circ}]$. Since the air-relative flight path angle is  $\gamma_a = \theta - \alpha$, assuming the pitch is well-regulated by the underlying autopilot and the angle of attack is well-regulated by the MPC algorithm, an air-relative flight path angle $|\gamma_a| = \pi/2$ is highly improbable and the NMPC problem will be feasible.

\subsection{MPCC Optimal Control Problem}

The form of the MPCC optimal control problem closely resembles (\ref{eq:mpc_ocp}), the main difference being the inclusion of the reference path parameter $\hat{\psi}$ and its derivative as decision variables within the optimization problem. To this end, define the MPCC aircraft state and control vectors as:

\begin{align} \label{eq:mpcc_state}
\begin{split}
    & \xb = [\rb^\top, \phi, \theta, \chi, V_a, \gamma_a ,  \delta_T, \Hat{\psi}]^\top \\
    & \ub = [\phi_c, \theta_c, \delta_{T,c}, \Dot{\psi}_c]^\top.
\end{split}
\end{align}

The aircraft dynamics (\ref{eq:aa_model}) are augmented with the reference path parameter dynamics: $\dot{\hat{\psi}} = \dot{\psi}_c$. We seek an MPCC controller which trades off path-following and path-progression metrics. To incentivize path progression, define the airspeed error term as:

\begin{equation}
    \epsilon_{V_a} = \Bar{V_{a}} - V_a.
\end{equation}

The stage error vector $\yb_k$ is augmented with $\epsilon_{V_a} = \Bar{V_{a}} - V_a$ to incentivize greater airspeed and therefore faster path progression. The reference path parameter $\hat{\psi}$ at stage $k=0$ is set to $\psi^*_i$ via (\ref{optimal_psi}) prior to solving the MPCC OCP. The MPCC OCP takes the exact same form as Equation \ref{eq:mpc_ocp}, where $R_k = \lambda^k \text{diag}([r_{\phi}, r_{\theta},r_{V_a}, r_{\dot{\psi}}]) \geq 0$. Weight matrix $Q$ remains a non-negative diagonal matrix augmented with $\mu$ to scale $\epsilon_{V_a}$ at stages with $k = [0, \dots, N-1]$. $\mu$ is tuned by the user to tradeoff path error with airspeed, with greater values favoring greater airspeed. At $k=N$, $\mu$ is set to $0$ to maximize the path error incentive at the final stage. Note that this formulation of MPCC is equivalent to weighting lag and contour errors equally \cite{liniger2015optimization, romero2022model}, a setting which was found to work well in practice for the FW-sUAS path-following problem. Finally, a hard box constraint is added to $\dot{\psi}_c$ (Table \ref{tab:mpc_params}) to incentivize a minimum rate of path progression.

\subsection{Implementation Details}

MPCC and CR-MPC were implemented using \texttt{acados} \cite{verschueren2022acados}, with \texttt{casadi} \cite{andersson2019casadi} used for automatic differentiation. OCPs were solved using sequential quadratic programming (SQP), with the real-time iteration (RTI) scheme for fast feedback \cite{gros2020linear}. The real-time iteration scheme solves a single iteration of SQP before returning a control signal, making a tradeoff of sub-optimality of the solution for faster feedback. By limiting SQP iterations, SQP-RTI also results in relatively consistent feedback delays. While not the focus of this paper, convergence and stability properties of SQP-RTI \cite{diehl2005real,liao2020time} and path-following MPC \cite{faulwasser2015nonlinear, diwale2017model} has been studied in other works. Due to the complexity of the aircraft dynamics and desired paths, the path-following MPC controllers in this work are formulated without provably stabilizing terminal costs and constraints. Stability via sufficiently long MPC horizons is empirically demonstrated for both controllers, as is common in practice \cite{grune2017nonlinear}. As the OCPs (\ref{eq:mpc_ocp}) are formulated in nonlinear least-squares form, the Gauss-Newton (GN) Hessian approximation can be readily utilized. The GN Hessian is positive semi-definite by definition, therefore the SQP-RTI subproblems are convex quadratic programs and can be solved efficiently \cite{rawlings2017model}. 

With each call of the MPC controllers, $\psi^*$ must be found relative to the latest aircraft positions. At initialization, the globally closest path parameter is found via a dense K-D tree query. For subsequent calls, a line search is conducted over a local region around the previous closest path parameter \cite{brito2019model}.

\begin{figure}
    \centering
    \includegraphics[width=\columnwidth]{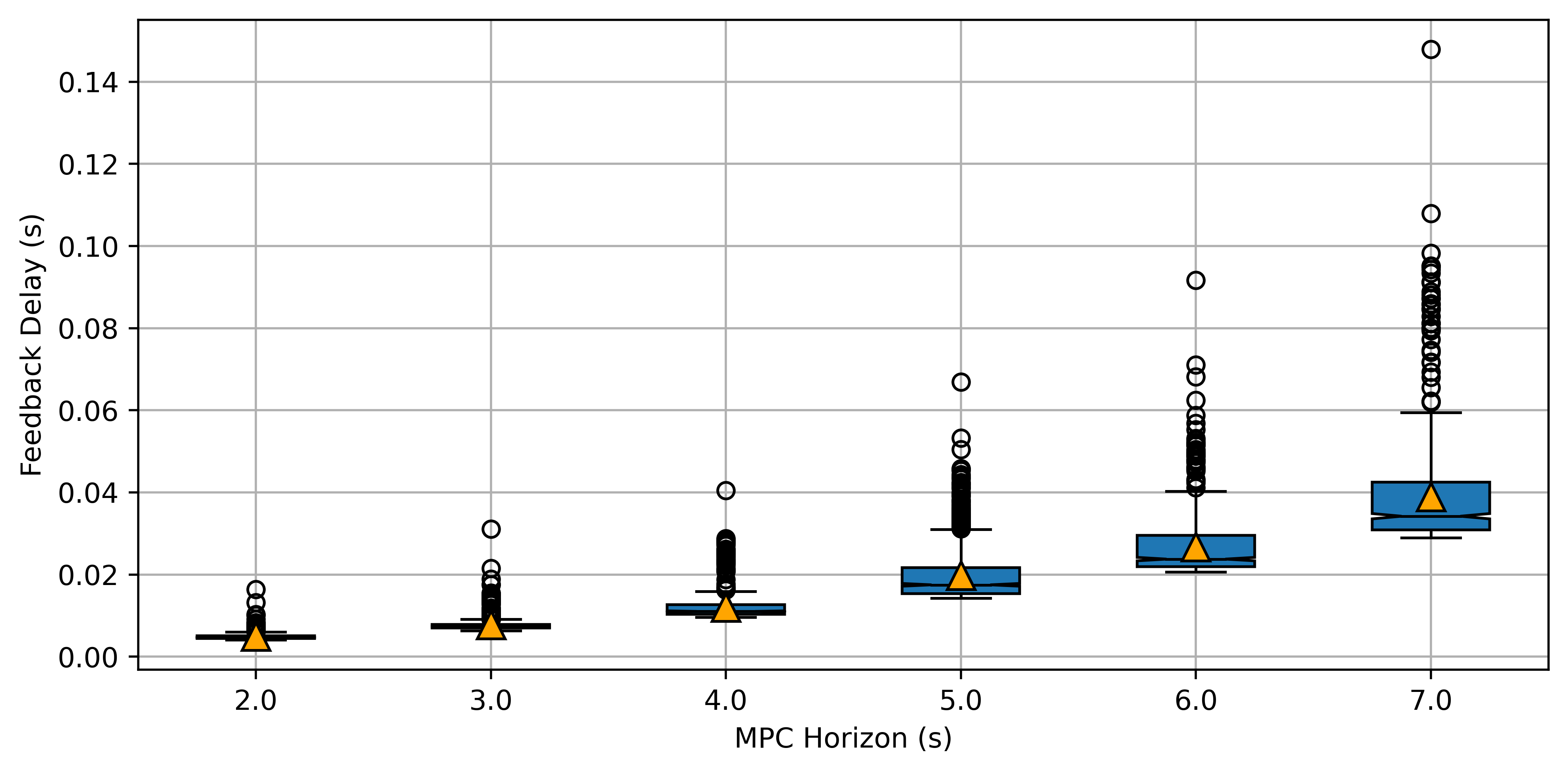}
    \caption{MPCC solve times over OCP horizon lengths during hardware-in-the-loop experiments. An OCP horizon of $T_f = 5$s was chosen to maximize closed-loop performance while maintaining low feedback delays for operation at 10 Hz.}
    \label{fig:solve_time_vs_horizon}
\end{figure}

Both MPC controllers used a time horizon $T_f = 5$ s, with constant discrete time interval $\Delta t = 0.1$ s over the horizon. Therefore, the number of stages $N = 50$. During flight experiments, the guidance controllers flown were run at 10 Hz. These settings were chosen to maximize performance while consistently achieving low feedback times under the onboard computational constraints (Figures \ref{fig:solve_time_vs_horizon}, \ref{fig:path_2_comp_times}). Tuning parameters for CR-MPC and MPCC can be found in Table \ref{tab:mpc_params}.

\begin{table}[]
\centering
\caption{MPC tuning parameters during flight experiments.}
\label{tab:mpc_params}
\resizebox{\columnwidth}{!}{%
\begin{tabular}{ll|ll}
Parameter                         & Value  & Parameter                & Value       \\ \hline
$q_n, q_e, q_d$                  & 1,1,1  & $r_{\phi}, r_{\theta}, r_{\delta_T}$ & 400,400,400 \\
$q_\chi, q_\gamma$                 & 1,1    & $r_{\Dot{\psi}}$                & 0.1         \\
$b_{\Dot{\phi}}, b_{\Dot{\theta}}, b_{\Dot{\delta_T}}$ & 1,20,10 & $s_{\alpha}, s_{V_a}$         & 1e4,1e4 \\
$\mu$                                & 0.001  & $\lambda$ & 0.99 \\
$\ubar{\phi}_c, \Bar{\phi}_c$                                & -45\degree, 45\degree  & $\ubar{\theta}_c, \Bar{\theta}_c$ & -10\degree, 10\degree \\
$\ubar{V}_a, \Bar{V}_a$                                & 20 m/s, 40 m/s  & $\ubar{\alpha}, \Bar{\alpha}$ & -6\degree, 12\degree \\
$\ubar{\delta}_{T,c}, \Bar{\delta}_{T,c}$                                & 0, 1  & $\ubar{\Dot{\psi}}_c, \Bar{\Dot{\psi}}_c$ & 15 m/s, 45 m/s
\end{tabular}%
}
\end{table}

\section{Flight Experiments}
\label{section:flight_experiments}

Flight experiments to test the CR-MPC and MPCC guidance algorithms were conducted in Arvada, Colorado, USA. Air temperature was 8\degree C, with a variable breeze (2-5 m/s) from the southeast. As a baseline comparison, the guidance controller in \cite{bird2014closing} was implemented. This is a version of lookahead guidance \cite{park2007performance, park2011autonomous}, a ubiquitous guidance law which is designed to follow paths of constant curvature (i.e. lines and circles). The lookahead controller was tuned in previous flight tests, and utilizes a lookahead time of 4 seconds while airspeed is regulated to a constant 21 m/s using a PID controller for throttle control. This relatively low reference airspeed was chosen to give the lookahead controller a better opportunity to stay on the challenging test paths.

\begin{figure}
    \centering
    \includegraphics[width=\columnwidth]{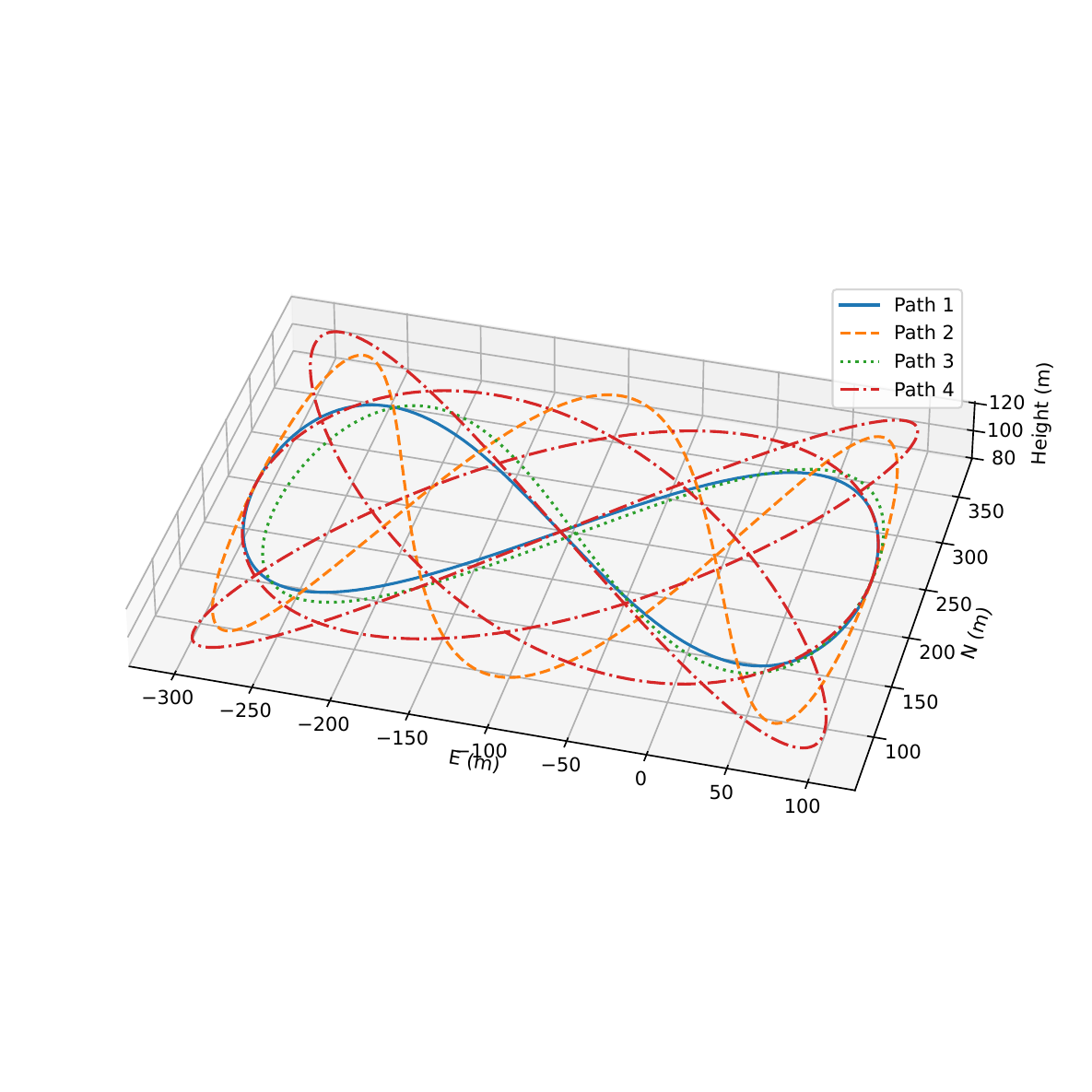}
    \caption{Four test paths flown during flight experiments. Paths 1 and 2 are at constant 100 m altitude, while paths 3 and 4 oscillate between 80 and 120 m altitude.}
    \label{fig:test_paths}
\end{figure}

\begin{figure}
    \centering
    \includegraphics[width=\columnwidth]{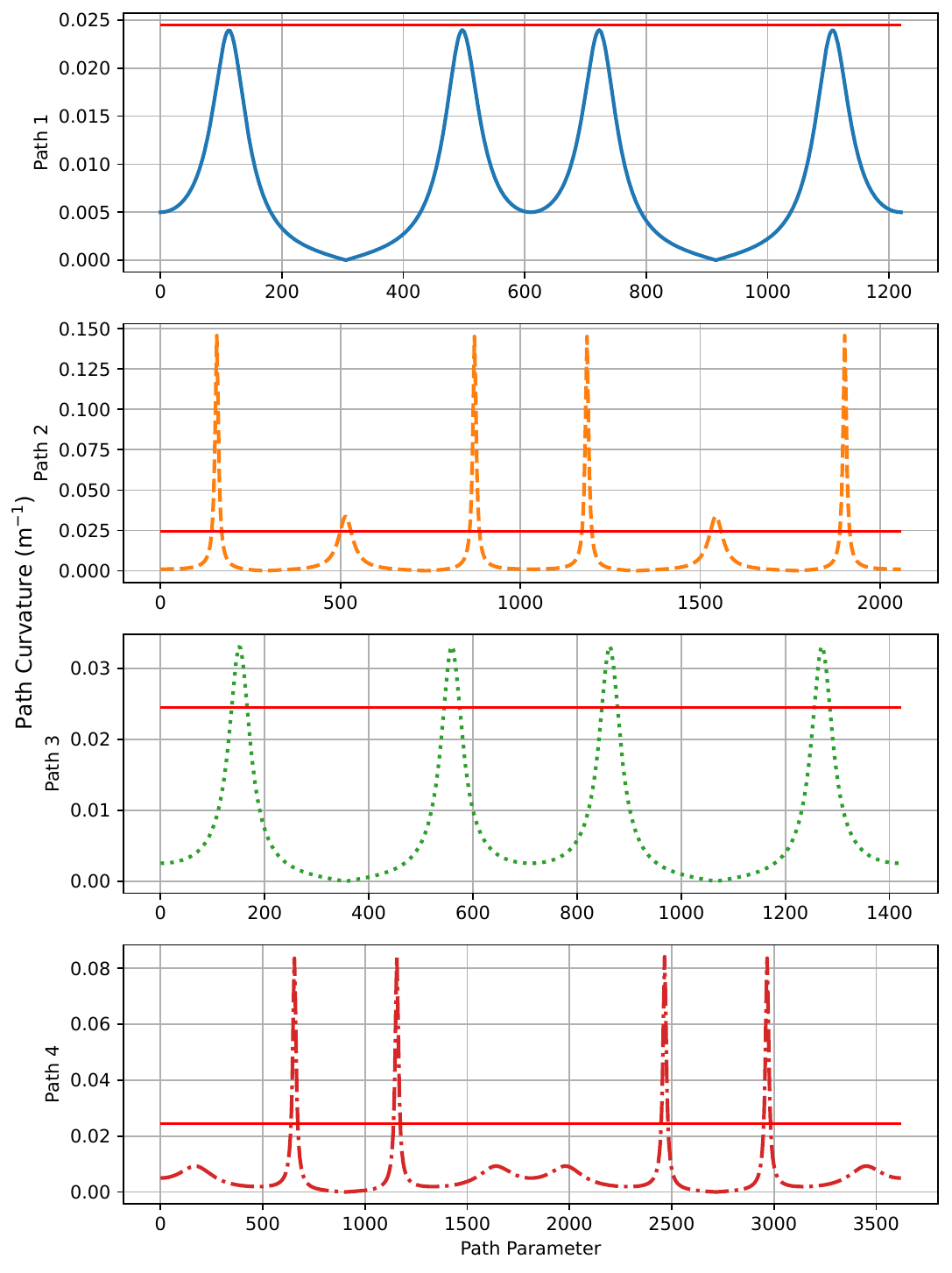}
    \caption{Path curvature of test paths, as a function of path parameter $\psi$. Red lines indicate the coordinated turn curvature of the RAAVEN at 20 m/s, 45\degree  roll.}
    \label{fig:path_curvature}
\end{figure}

All guidance controllers were run onboard an NVIDIA Jetson Xavier NX companion computer which communicates with the Cube Orange flight computer via a wired serial connection. Onboard computation ensures minimal latency and data loss, and demonstrates the feasibility of all guidance algorithms for operation in communication constrained scenarios. Aircraft control and state data to and from the autopilot was streamed using the microRTPS and ROS2 middlewares. 

\input{performance_metric_table}

Four Lissajous curves were chosen as paths to test the strengths and weaknesses of each controller (Figure \ref{fig:test_paths}). Paths 1 and 2 are at a constant height of 100 m. Paths 3 and 4 oscillate vertically between 80 m and 120 m over the course of a lap. The coordinated turning radius of the RAAVEN in no wind is 40.8 m at a minimum airspeed of 20 m/s and maximum roll of 45\degree. Path 1 has a minimum curvature radius of 41.7 m, making it the only path with a minimum curvature radius larger than the coordinated turning radius of the RAAVEN. Paths 2, 3, and 4 are more challenging, with a minimum curvature radius of 6.9 m, 30.2 m, and 11.9 m respectively (Figure \ref{fig:path_curvature}). Path 3 is particularly challenging in the longitudinal axis, demanding a maximum flight path angle of 8.4\degree. It should be noted that all four paths may or may not be feasible to fly, depending on wind conditions. Paths were limited in horizontal size and altitude due to restrictions of the flight testing area. For each path, each guidance algorithm was flown for two laps, one after the other in an attempt to minimize effects from changing wind conditions. Efforts were made to initialize the aircraft similarly for each two-lap episode. For CR-MPC, the reference path rate was fixed at 25 m/s for all experiments. Other MPC controller parameters (Table \ref{tab:mpc_params}) were held constant across all paths.

Summary statistics for path-following error, airspeed, ground speed, and feedback time for each controller and path are presented in Table \ref{tab:performance_metrics}. 

\begin{figure}
    \centering
    \includegraphics[width=\columnwidth]{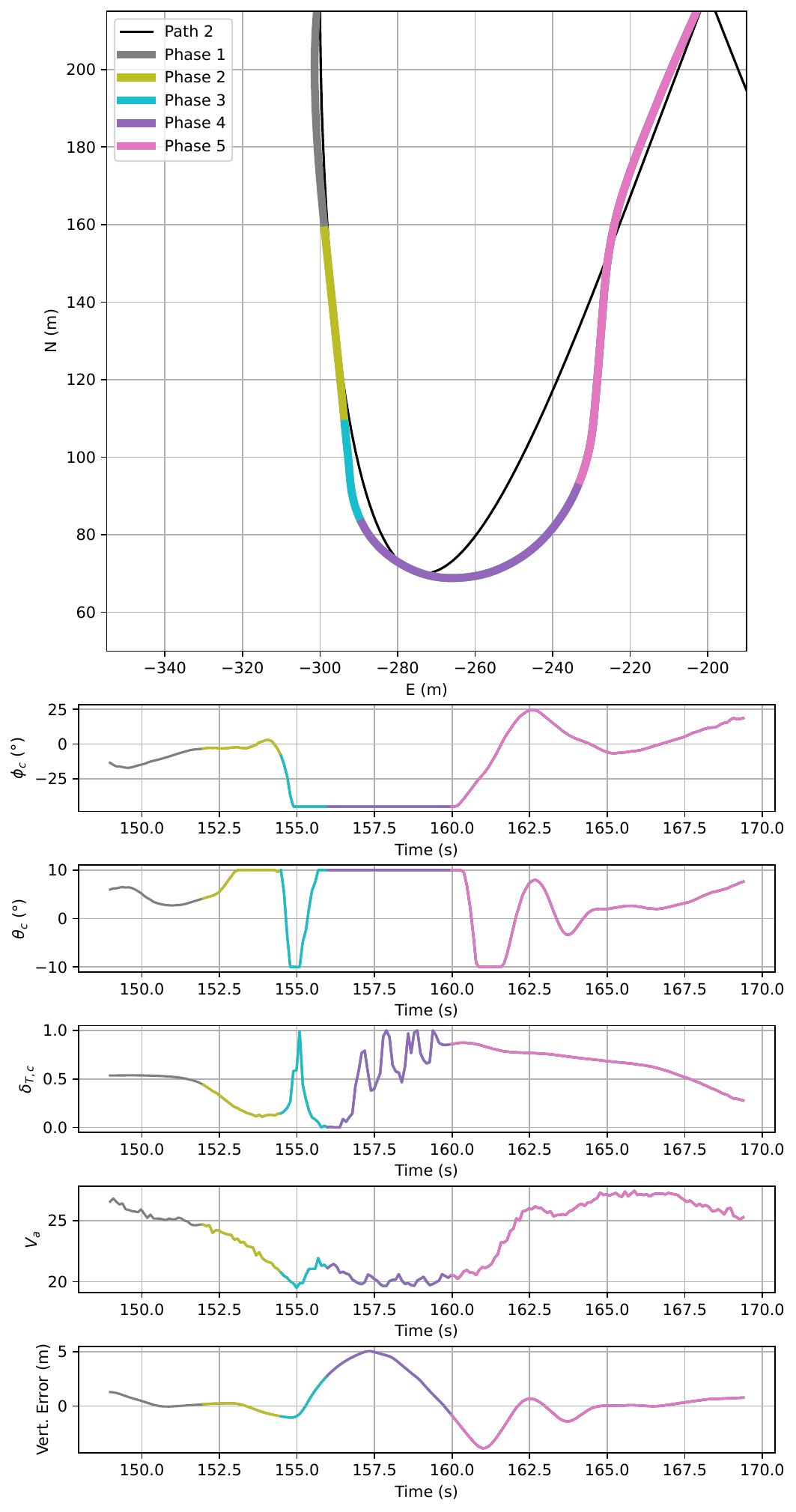}
    \caption{Phases of a tight turn maneuver exhibited by both MPCC and CR-MPC over path 2. Only MPCC data is visualized for brevity. Phase 1 is approaching the turn on path, at high airspeed and with level wings. Phase 2 cuts throttle and pitches up to reduce airspeed upon close approach to the turn. Phase 3 initiates the turn by banking hard and momentarily pitching down to keep airspeed admissible and reduce altitude in preparation for the next phase. Phase 4 commands maximum roll and pitch up to minimize the turning radius of the aircraft, while throttling just enough to keep the airspeed admissible. Phase 5 maneuvers the aircraft to get back on the path, completing the turn. These complex behaviors are not explicitly encoded, but are found implicitly by solving the OCP online. Standard guidance methods are unable to conduct maneuvers such as these.}
    \label{fig:mpcc_tight_turn_phases}
\end{figure}

\begin{figure*}
     \centering
     \begin{subfigure}[b]{0.49\textwidth}
         \centering
         \includegraphics[width=\textwidth]{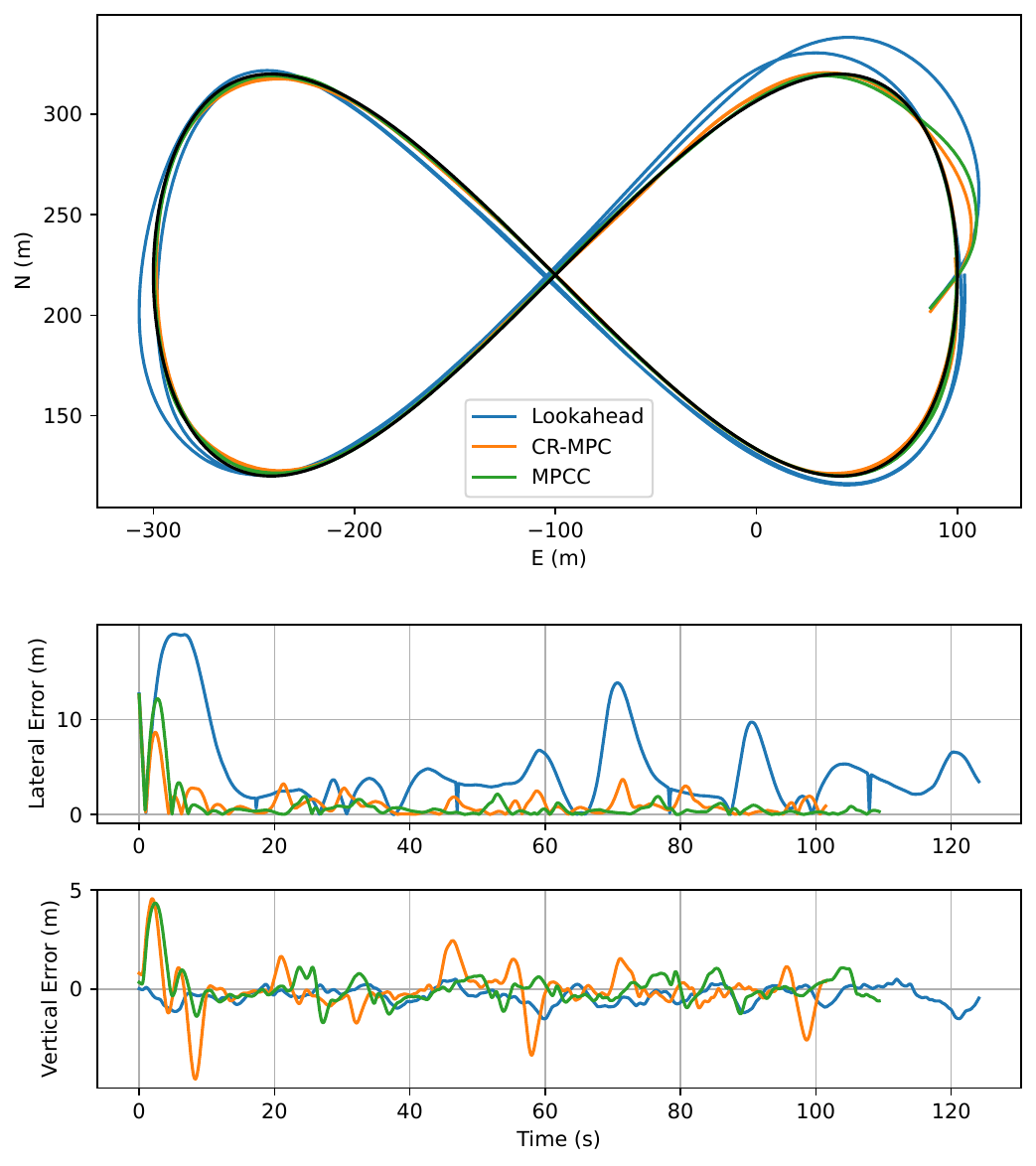}
         \caption{Path 1}
         \label{fig:path_1_path_error}
     \end{subfigure}
     \hfill
     \begin{subfigure}[b]{0.49\textwidth}
         \centering
         \includegraphics[width=\textwidth]{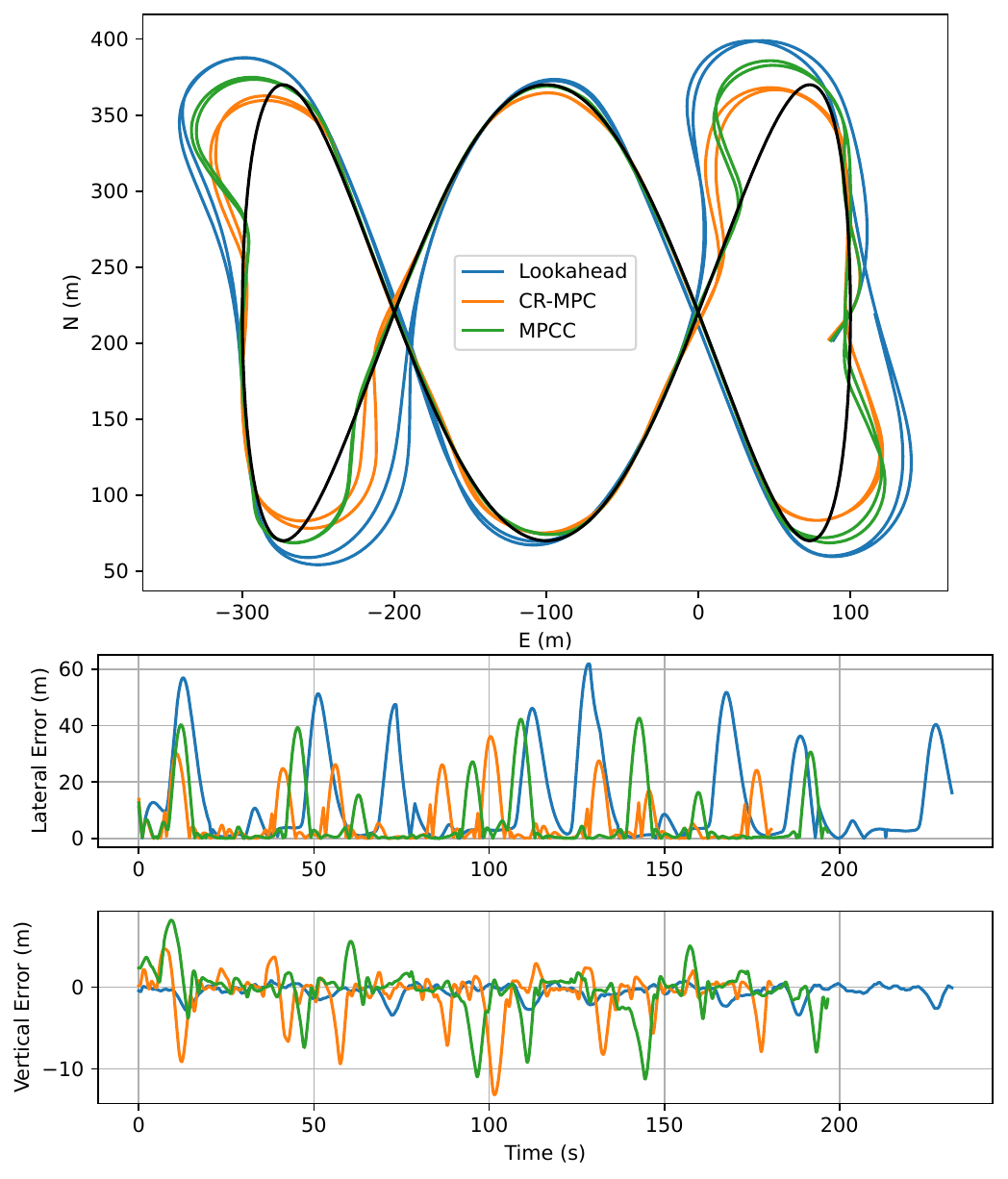}
         \caption{Path 2}
         \label{fig:path_2_path_error}
     \end{subfigure}
     \hfill
     \begin{subfigure}[b]{0.49\textwidth}
         \centering
         \includegraphics[width=\textwidth]{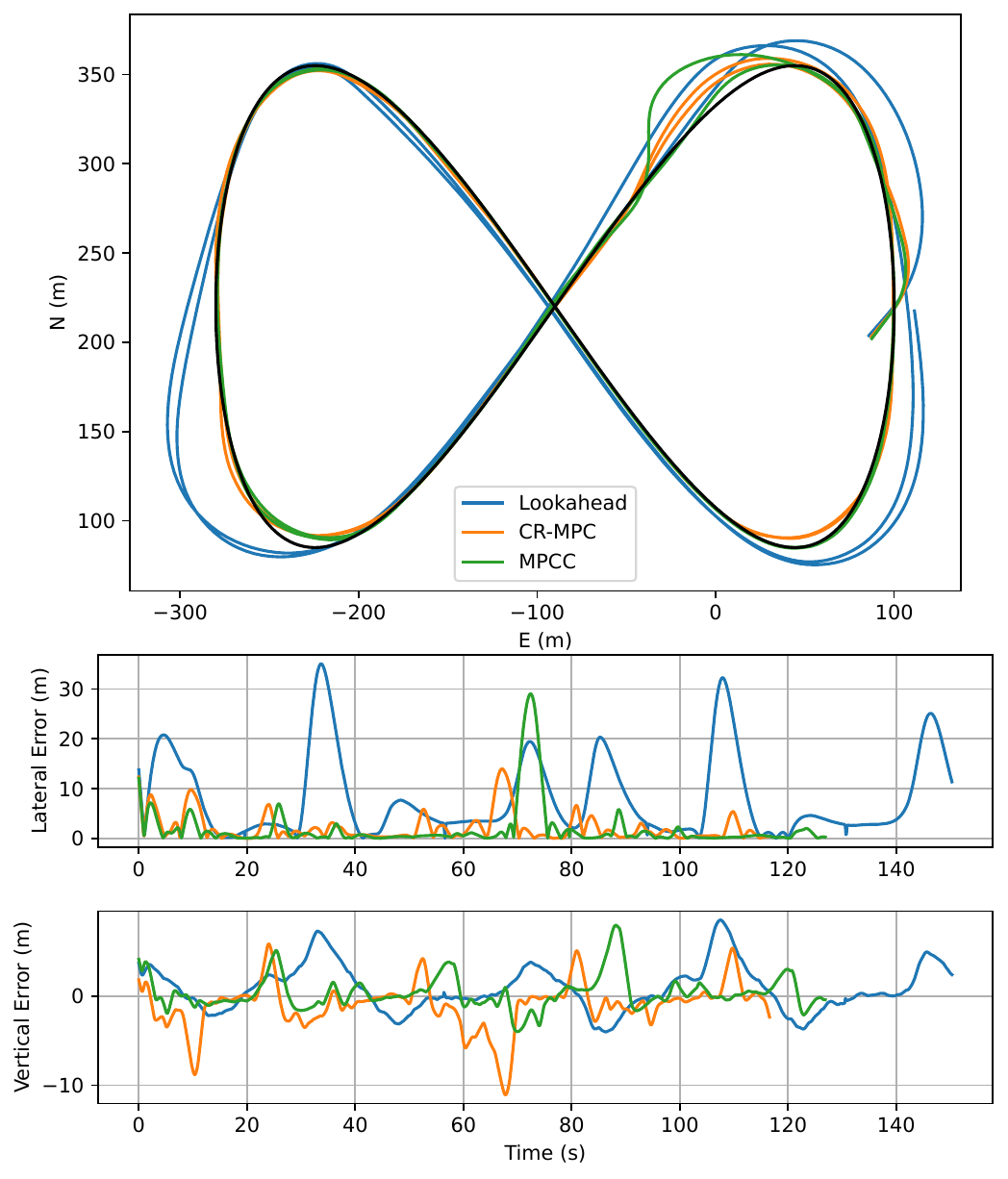}
         \caption{Path 3}
         \label{fig:path_3_path_error}
     \end{subfigure}
     \hfill
     \begin{subfigure}[b]{0.49\textwidth}
         \centering
         \includegraphics[width=\textwidth]{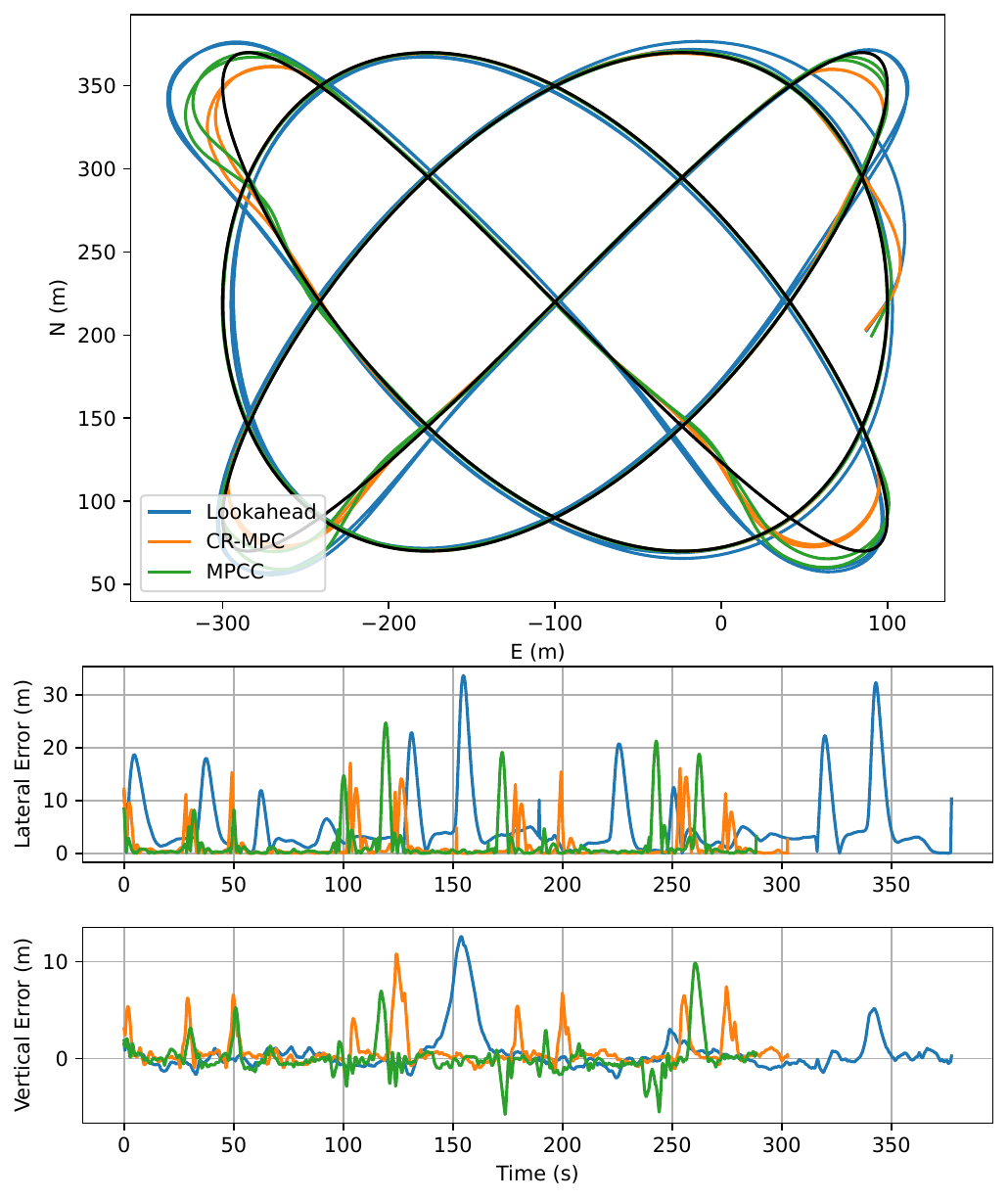}
         \caption{Path 4 }
         \label{fig:path_4_path_error}
     \end{subfigure}
        \caption{Aircraft lateral position history, lateral path error, and vertical path error of each guidance algorithm while running over two laps of each test path.}
        \label{fig:path_error}
\end{figure*}

\begin{figure*}
     \centering
     \begin{subfigure}[b]{0.49\textwidth}
         \centering
         \includegraphics[width=\textwidth]{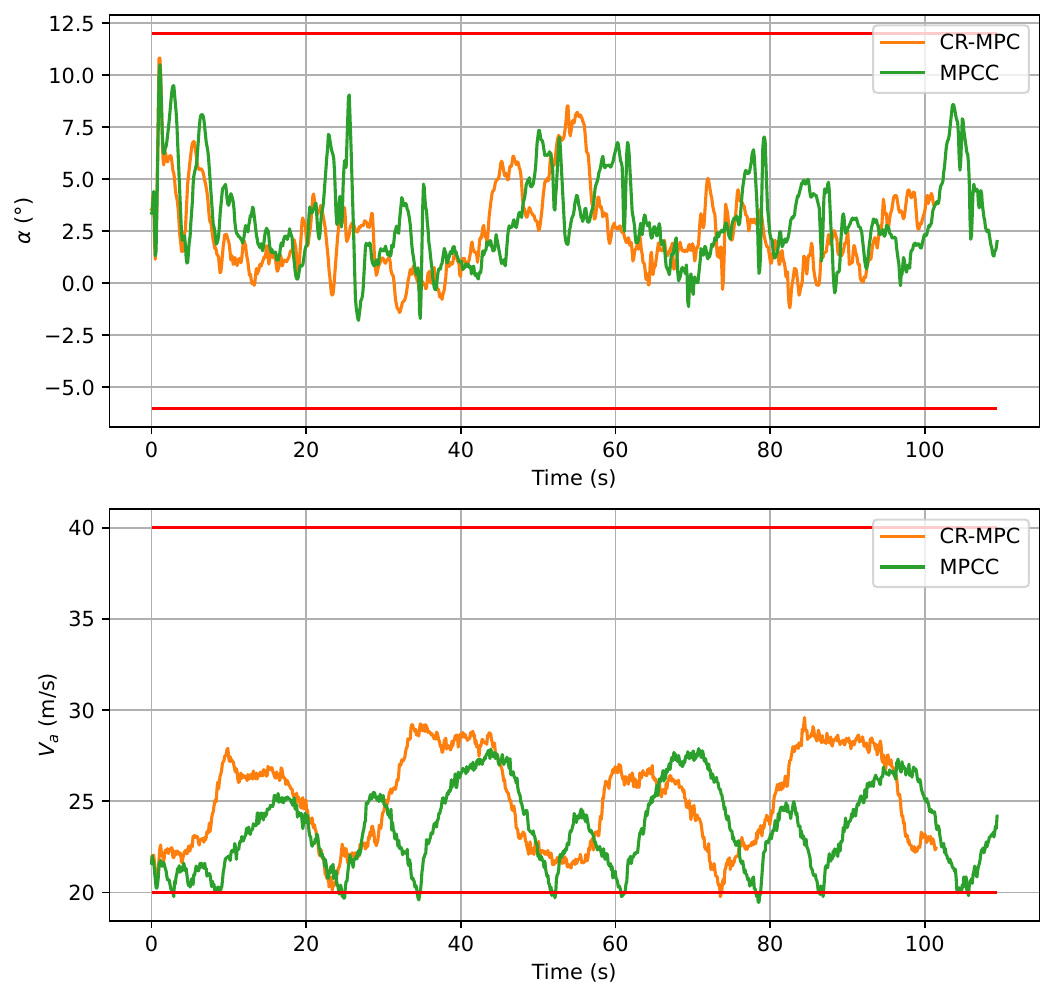}
         \caption{Path 1}
         \label{fig:path_1_soft_constr}
     \end{subfigure}
     \hfill
     \begin{subfigure}[b]{0.49\textwidth}
         \centering
         \includegraphics[width=\textwidth]{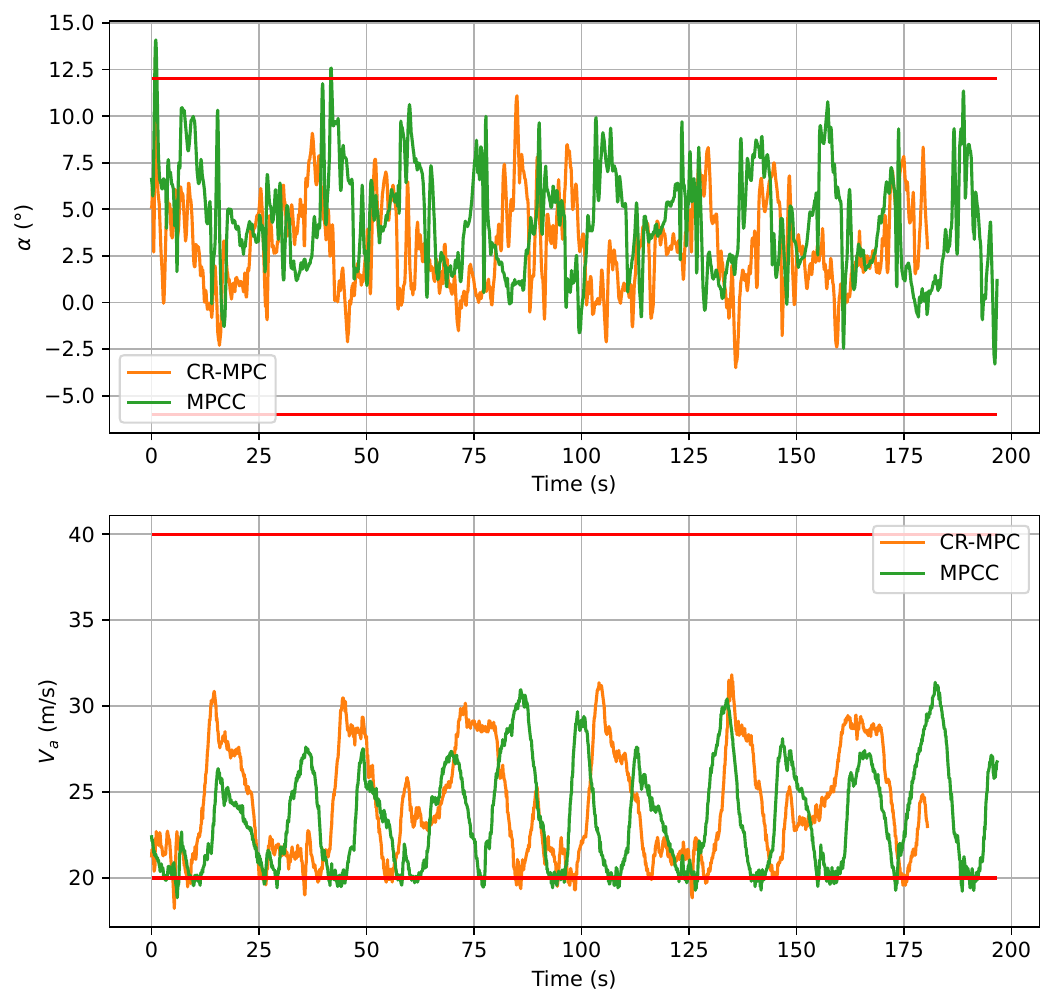}
         \caption{Path 2}
         \label{fig:path_2_soft_constr}
     \end{subfigure}
     \hfill
     \begin{subfigure}[b]{0.49\textwidth}
         \centering
         \includegraphics[width=\textwidth]{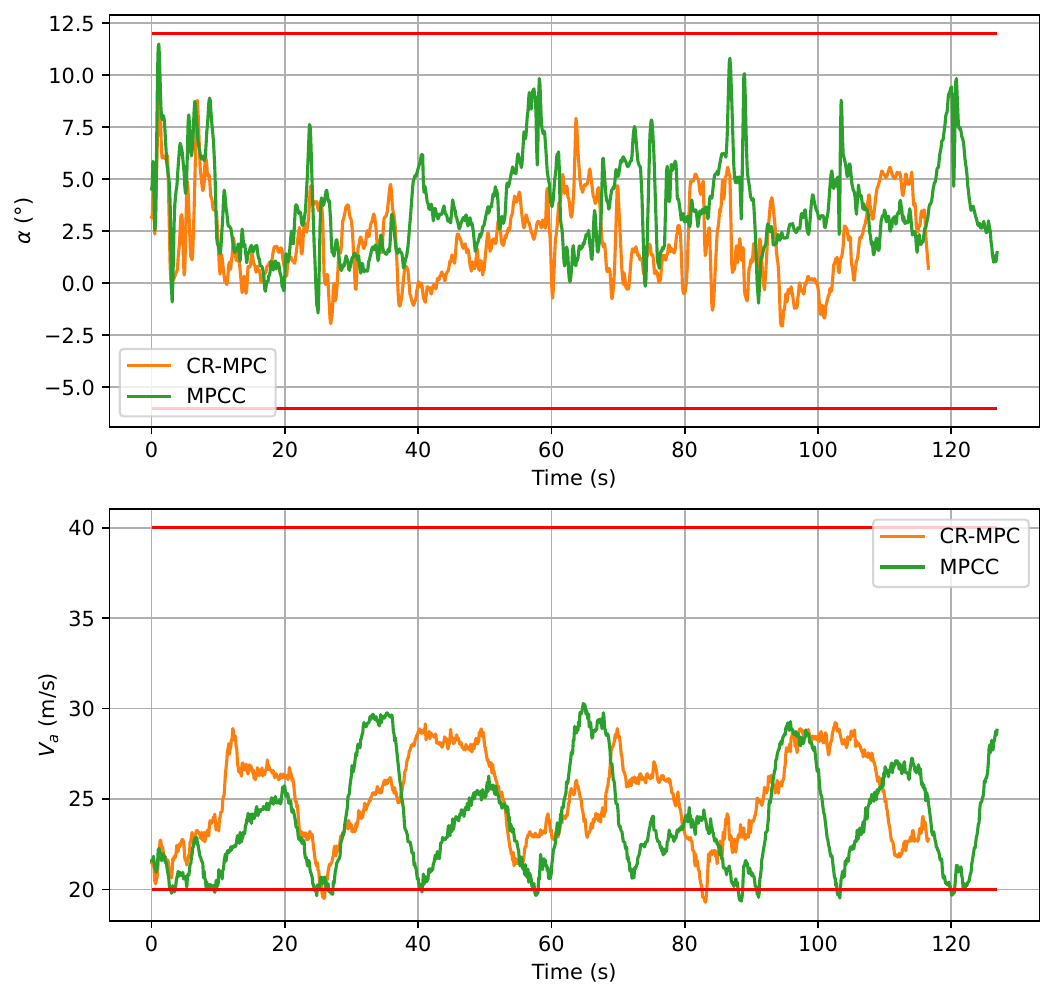}
         \caption{Path 3}
         \label{fig:path_3_soft_constr}
     \end{subfigure}
     \hfill
     \begin{subfigure}[b]{0.49\textwidth}
         \centering
         \includegraphics[width=\textwidth]{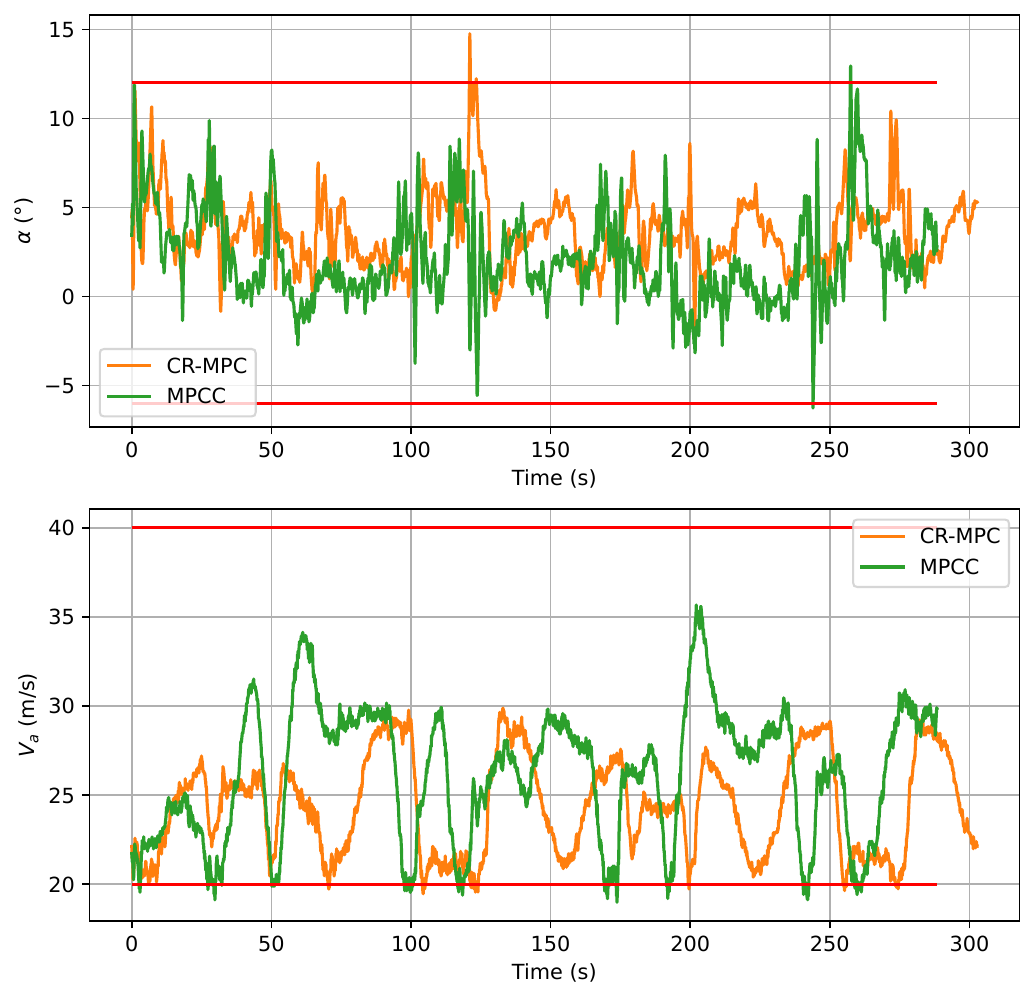}
         \caption{Path 4}
         \label{fig:path_4_soft_constr}
     \end{subfigure}
        \caption{Angle of attack and airspeed histories for each MPC controller over two laps of each test path.}
        \label{fig:soft_constr}
\end{figure*}

Across all paths, path-following error (i.e. distance from path) of the MPC-based algorithms was observed to be much lower than the lookahead controller. This is readily attributed to the MPC controllers' total control authority over pitch/throttle, and therefore airspeed. When combined with a suitably long OCP horizon, MPC-based controllers proactively reduce airspeeds when approaching tight turns to reduce the turning radius of the aircraft (Figure \ref{fig:soft_constr}). For particularly tight turns, as with path 2, both MPC controllers were observed to conduct a complex sequence of maneuvers in roll, pitch, and throttle that effectively further reduce the turning radius of the aircraft while maintaining low path errors (Figure \ref{fig:mpcc_tight_turn_phases}). Both MPC algorithms achieved low path-following error while simultaneously flying at consistently higher air and ground speeds. In contrast, the lookahead controller is myopic and cannot effectively regulate airspeed to achieve high-performance path following. The stark contrast between MPC and lookahead guidance is visualized in Figure \ref{fig:path_error}. 

It was also observed that MPC controllers exhibit greatly reduced lateral path error, but higher vertical path error than lookahead guidance across paths 1, 2, and 4 (Figure \ref{fig:path_error}). This is attributed to the MPC cost functions, which weight errors in the lateral and vertical errors equally (Table \ref{tab:mpc_params}). This is equivalent to minimizing the absolute distance from path, which is sometimes accomplished via maneuvers which induce vertical errors. Increasing the cost weight of vertical path error ($q_d$) would incentivize MPC to maintain altitude, likely at the cost of greater lateral path error. The exception to this lies in path 3 (Figure \ref{fig:path_3_path_error}), which demands aggressive longitudinal maneuvering to maintain the desired altitude. MPC controllers were relatively more successful in this regard, as they proactively maneuvered pitch and throttle to meet the demanding vertical reference changes.

Discerning performance superiority between MPCC and CR-MPC is more challenging. Across paths, mean and median path-following error, ground speed, and airspeed statistics are all nearly identical between the two controllers. Changing and gusty wind conditions mean small statistical differences may simply not be representative of the controllers' true relative performance. 

\begin{figure}
    \centering
    \includegraphics[width=\columnwidth]{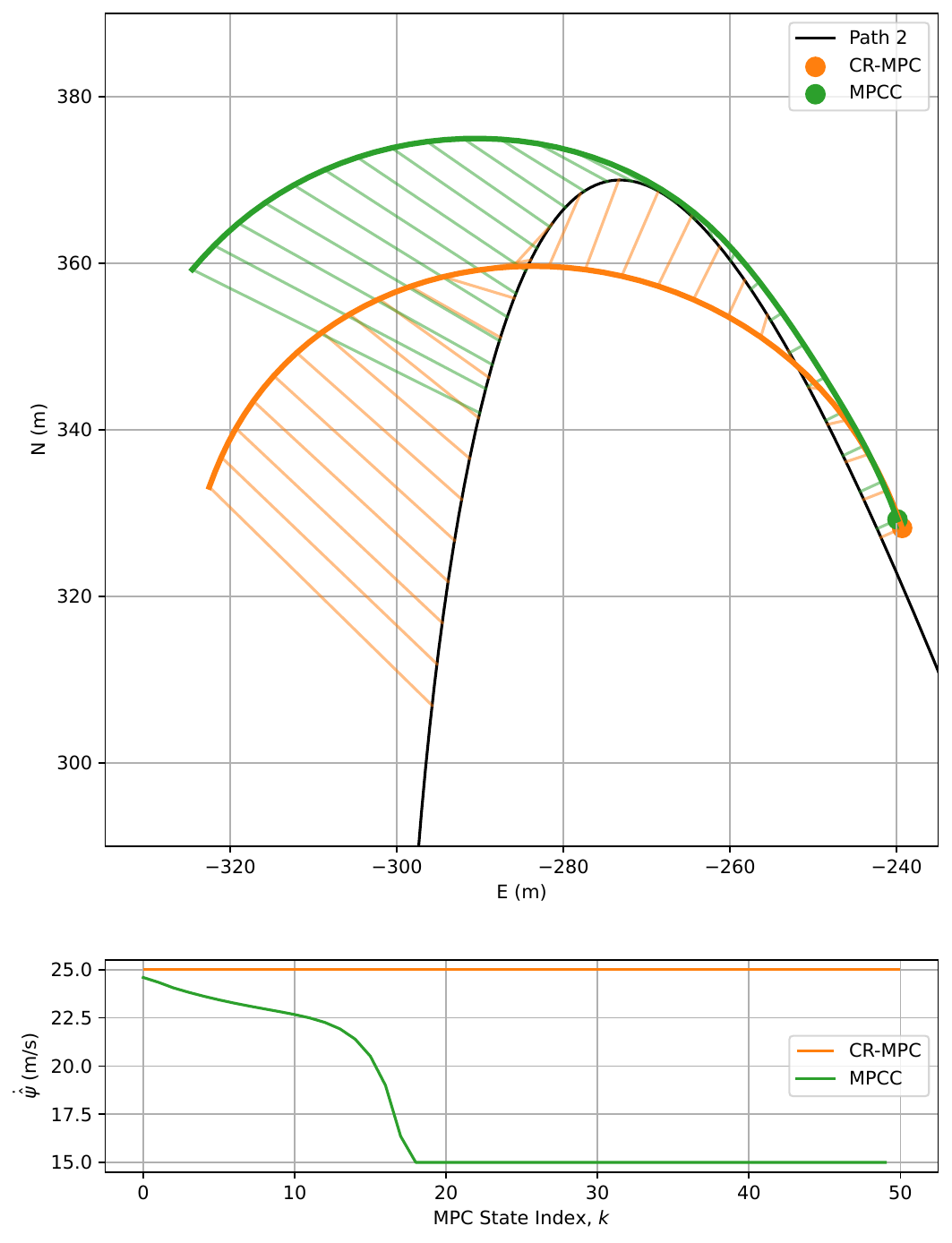}
    \caption{ [Top] Predicted aircraft positions over the MPC horizon as the aircraft (green and orange circles) approach a tight turn while flying test path 2. Bolded lines indicate the predicted positions over the MPC horizon. Thin lines are the lateral position error vectors $[\epsilon_N, \epsilon_E]^\top$ at every other stage in the MPC horizon. [Bottom] Values of path reference rate $\Dot{\hat{\psi}}$ over the MPC horizon for each controller. While MPCC reduces path-following error over the horizon, beyond the horizon the corner-cutting strategy exhibited by CR-MPC reduces path-following error beyond the horizon.}
    \label{fig:horizon_zoomed}
\end{figure}

One observed difference between CR-MPC and MPCC was that the constant $\Dot{\hat{\psi}}$ in CR-MPC incentivizes ``corner cutting", which is detrimental to path error on paths 1 and 3, but useful on the particularly tight turns on paths 2 and 4. Corner-cutting increases path-following error when paths can become feasible through slight reductions in ground speed, but can reduce path error on paths which are infeasible to follow. For example, on path 2, MPCC attempts to stay close to the path as long as possible, which increases overshoot and therefore path-following errors beyond the OCP horizon (Figure \ref{fig:horizon_zoomed}). Thanks to corner cutting, CR-MPC demonstrated lower maximum path-following error across all four paths. With a longer OCP horizon (and more computational effort), it is expected that MPCC could close the performance gap in these scenarios.

\begin{figure*}
    \centering
    \includegraphics[width=\textwidth]{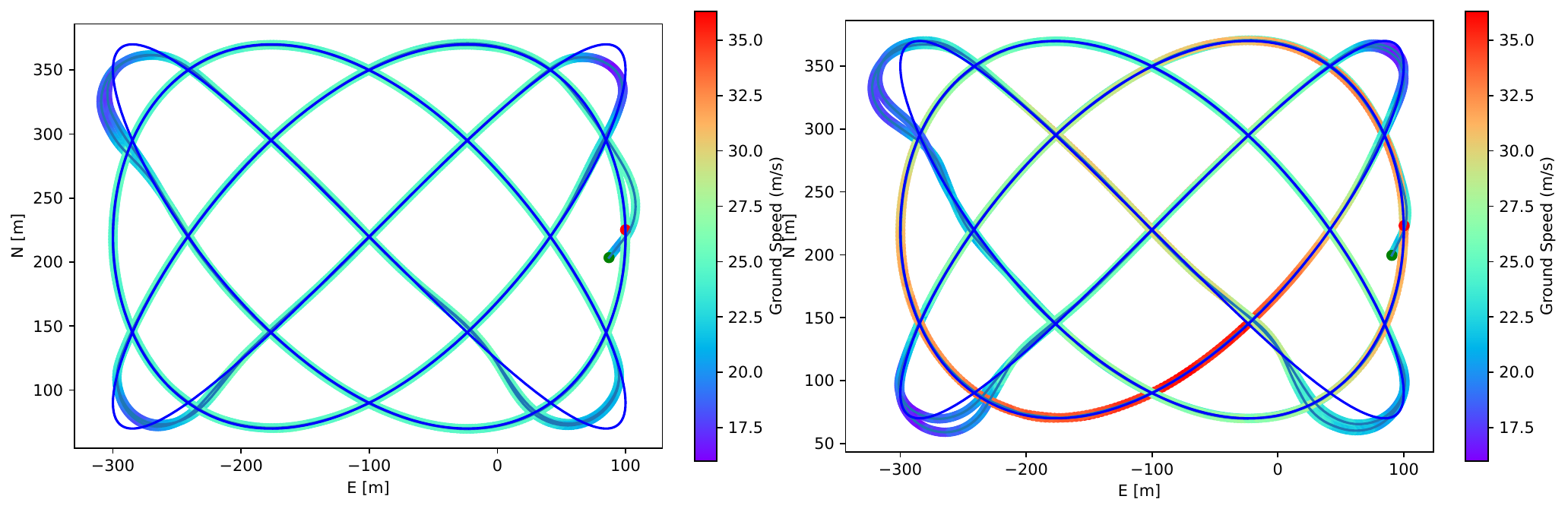}
    \caption{Ground speed of CR-MPC (left) and MPCC (right) while flying path 4. Green circle is the starting location, red circle is the final location after flying two laps. MPCC adapts its airspeed and ground speed to the local path segment, while CR-MPC maintains a near-constant ground speed at all times.}
    \label{fig:mpc_groundspeed_colorbar}
\end{figure*}

Other differences between MPCC and CR-MPC also arise when looking at the overall behavior of the controllers during flight. Figure \ref{fig:mpc_groundspeed_colorbar} showcases the effects of allowing MPCC to choose the path reference rate, which leads to a much wider range of ground speeds over path 4. CR-MPC maintains a much narrower range of ground speeds, as expected by setting a constant $\Dot{\hat{\psi}}$. MPCC induces a large range of ground speeds by more aggresively varying the airspeed, and therefore MPCC always has larger maximum ground speed than CR-MPC (Table \ref{tab:performance_metrics}). While both controllers result in high-performance path following, each may be better suited for different purposes. CR-MPC is suitable for applications which benefit from consistent ground speeds, such as large-scale surveying or mapping. MPCC is suitable for applications where a constant ground speed may easily become unattainable due to strong and dynamic wind conditions, such as severe storm research.

\begin{figure*}
     \centering
     \begin{subfigure}[b]{0.49\textwidth}
         \centering
         \includegraphics[width=\textwidth]{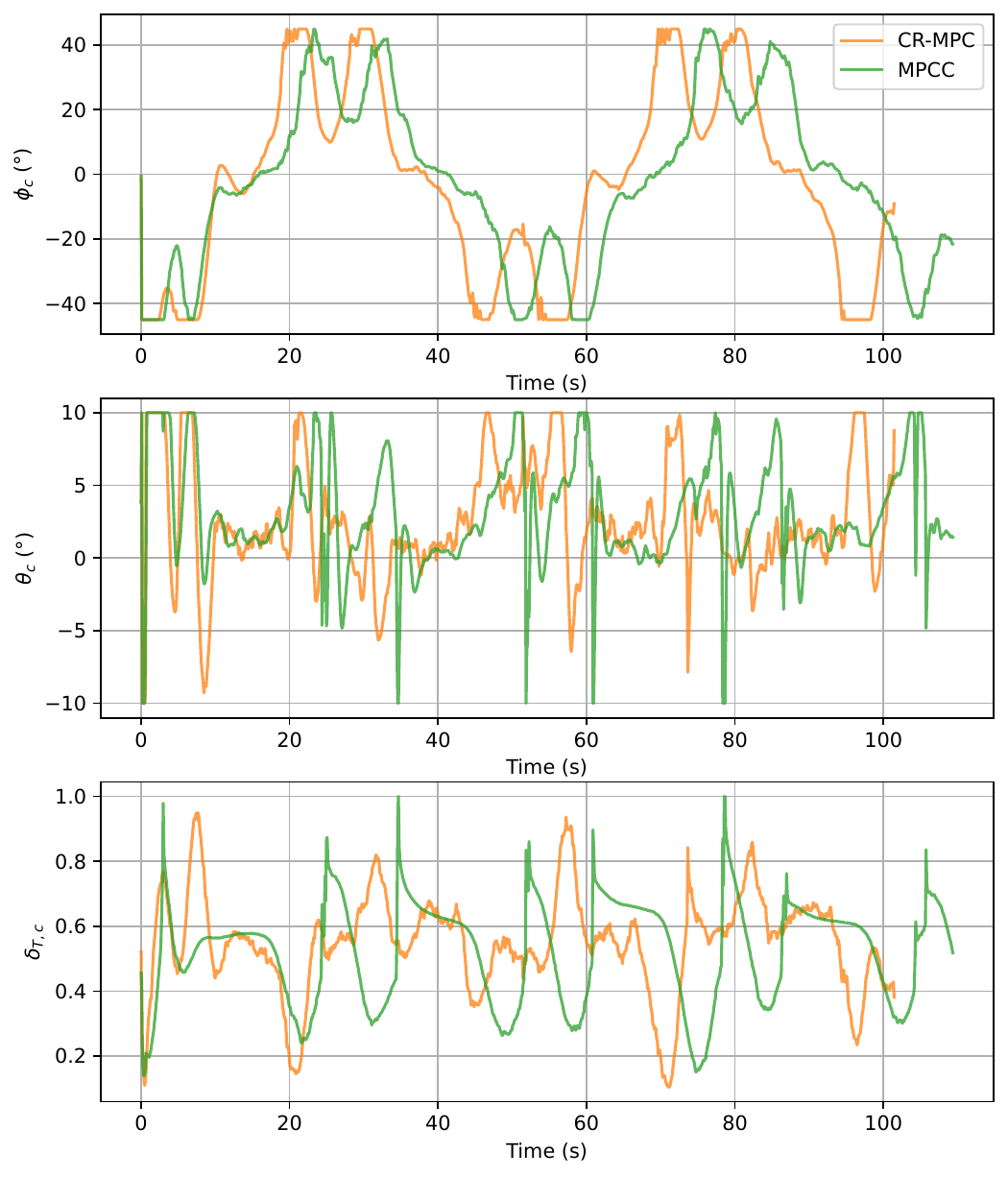}
         \caption{Path 1}
         \label{fig:path_1_controls}
     \end{subfigure}
     \hfill
     \begin{subfigure}[b]{0.49\textwidth}
         \centering
         \includegraphics[width=\textwidth]{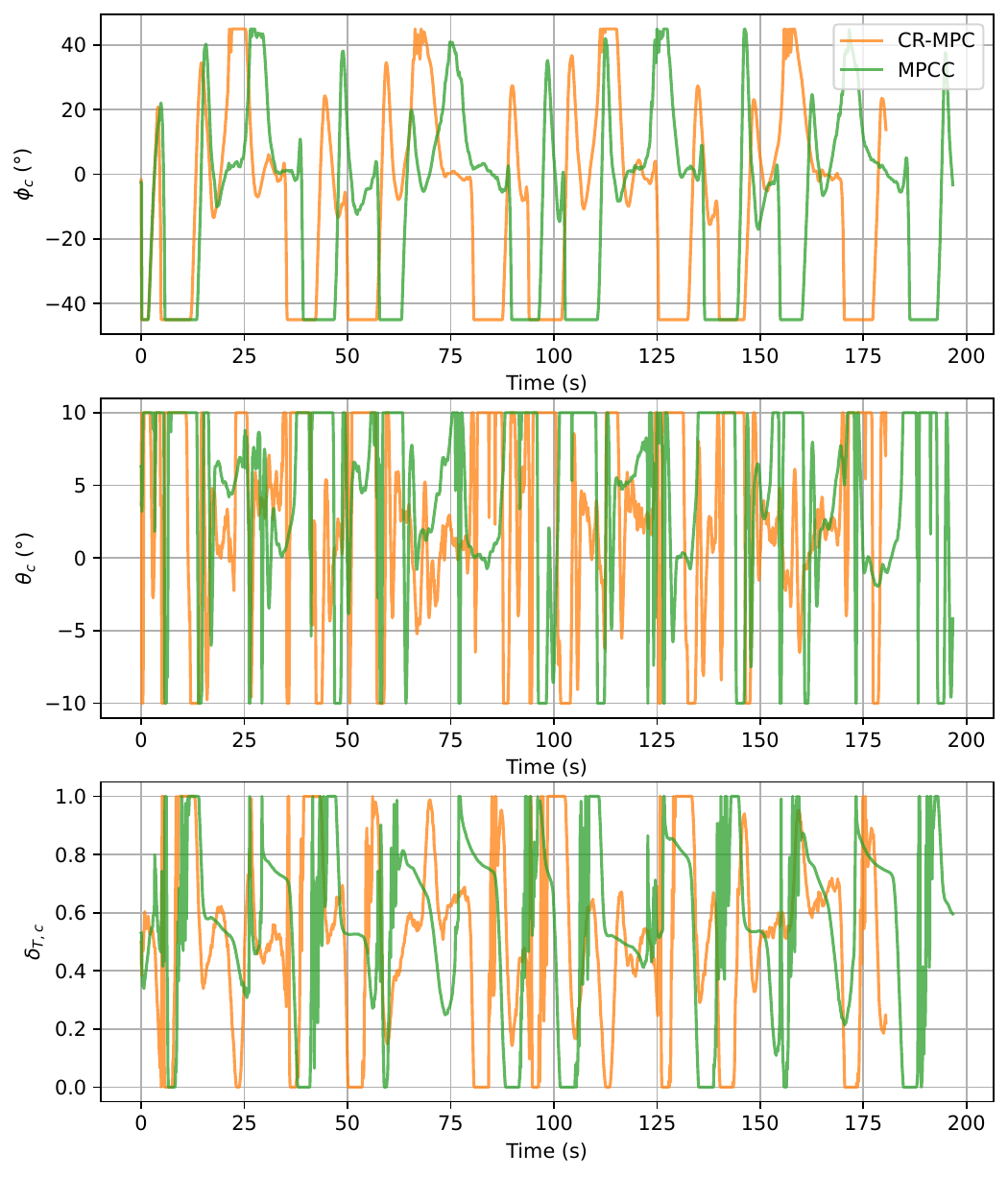}
         \caption{Path 2}
         \label{fig:path_2_controls}
     \end{subfigure}
     \hfill
     \begin{subfigure}[b]{0.49\textwidth}
         \centering
         \includegraphics[width=\textwidth]{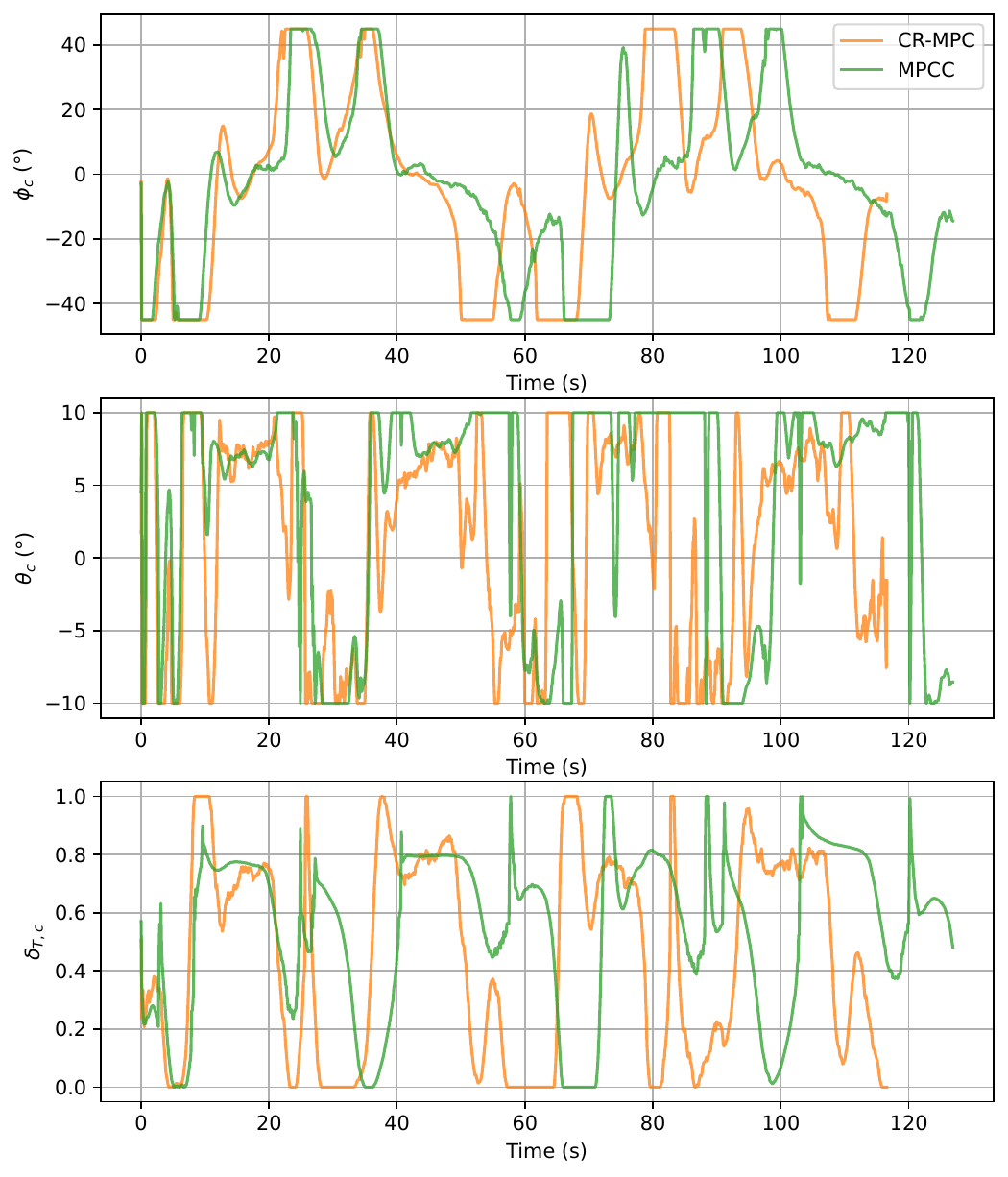}
         \caption{Path 3}
         \label{fig:path_3_controls}
     \end{subfigure}
     \hfill
     \begin{subfigure}[b]{0.49\textwidth}
         \centering
         \includegraphics[width=\textwidth]{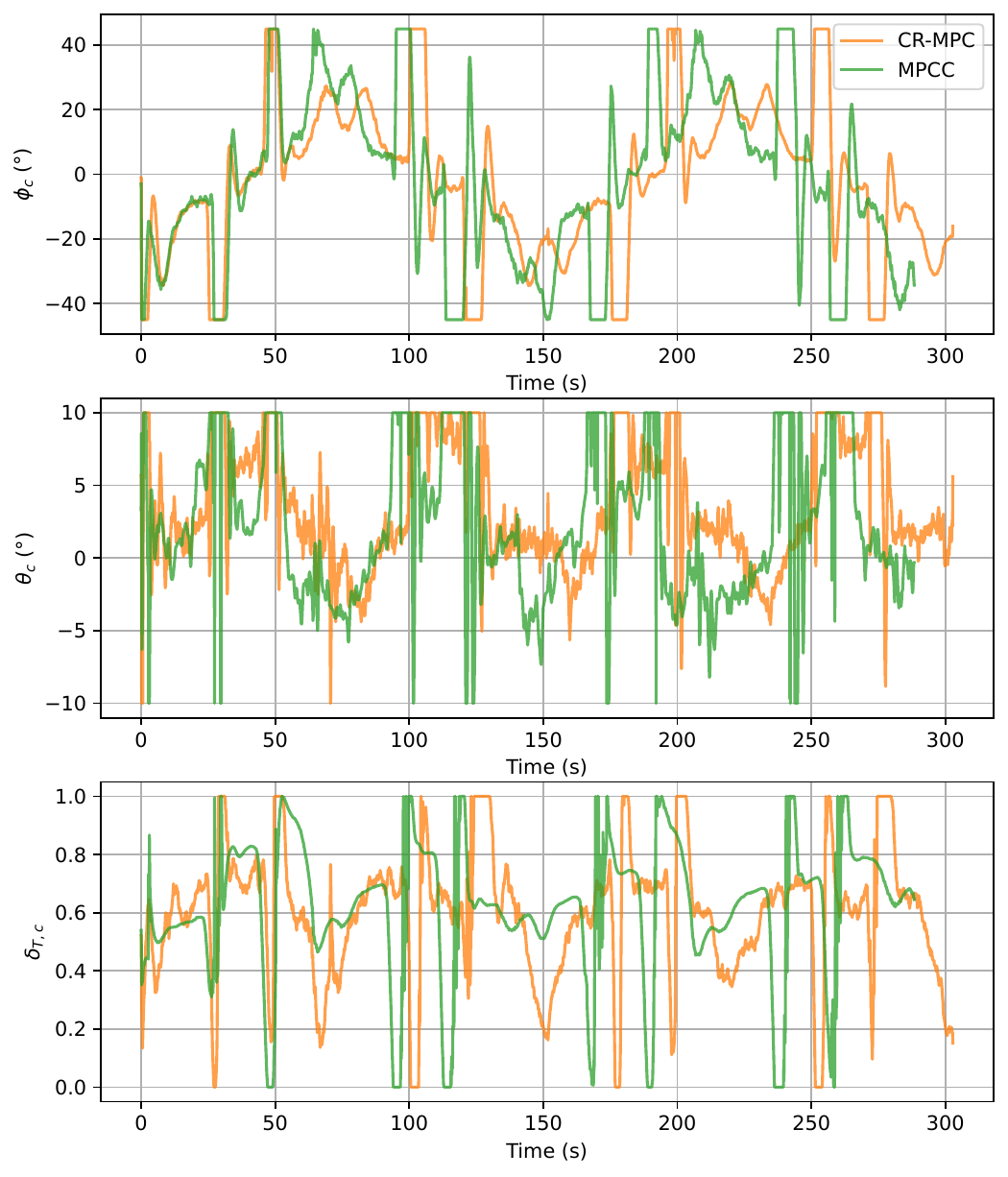}
         \caption{Path 4}
         \label{fig:path_4_controls}
     \end{subfigure}
        \caption{Commanded roll, pitch, and throttle histories for each MPC controller over two laps of each test path.}
        \label{fig:controls}
\end{figure*}

Control histories of CR-MPC and MPCC are shown in Figure \ref{fig:controls}. It is easily observed that pitch and throttle command signals are more ``bang-bang" than the roll signal, for both MPC algorithms. This can be partially attributed to the challenging longitudinal control problem, which deals with noisy measurements (i.e. airspeed, angle of attack), a constrained state space, and greater plant-model mismatch (Table \ref{tab:validation_rmse}). The bang-bang control is more common over the challenging paths 2 and 4, where the aircraft spends more time on the limits of the safe flight envelope to conduct tight turn maneuvers (Figure \ref{fig:soft_constr}). However, it is noted that the professional pilot-in-command qualitatively found the resulting flight behavior of all guidance controllers on all paths nominal and safe, never initiating a manual override.

Despite the challenges both MPC controllers face regarding sensor noise and plant-model mismatch, the MPC algorithms were able to keep airspeed and angle of attack largely within the pre-defined safe flight envelope (Figure \ref{fig:soft_constr}) via soft constraints (Equation \ref{eq:soft_constr}). The challenging characteristics of paths 2 and 4 mean more aggressive maneuvers, and therefore more time spent near the boundaries of the flight envelope.

\begin{figure}
    \centering
    \includegraphics[width=\columnwidth]{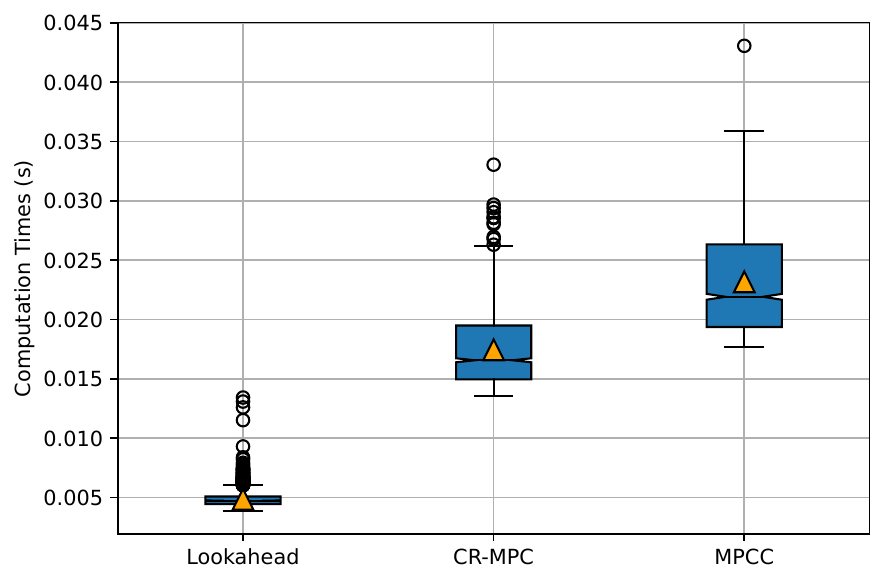}
    \caption{Computation time distributions for each controller over Path 2 during flight testing. These times are equivalent to feedback delay, i.e. time from when the controller is queried to when a guidance command is sent to the autopilot. MPC computation times are more variable than lookahead guidance due to the inclusion of the iterative OCP solver.}
    \label{fig:path_2_comp_times}
\end{figure}

Finally, it was shown that CR-MPC is more computationally efficient than MPCC, with mean feedback times lower by over 20\% across all paths. This computational savings is simply a result of the increased number of decision variables within the MPCC OCP. Despite their computational cost, both CR-MPC and MPCC are run well within the 10Hz threshold, even over the most difficult path and longest feedback times (Figure \ref{fig:path_2_comp_times}).

\section{Discussion}
\label{section:discussion}

The flight experiment results in Section \ref{section:flight_experiments} clearly demonstrate the utility and feasibility of the developed MPC algorithms. Fast and precise 3D path-following guidance will prove useful across a wide range of applications. Along with these positive results, there are several opportunities for future work.

First, it is again noted that wind measurements were taken directly from the PX4 extended kalman filter, which only produces measurements of the lateral wind components $w_n, w_e$. This was done to showcase how MPC could be used with PX4, without modification, to conduct system identification and path-following guidance. Therefore, the vertical wind component was assumed $w_d = 0$ throughout the paper, which allowed for calculations of the air-relative velocity vector, and therefore $\alpha$. Improved sensing of the 3D wind vector, either through multihole probes or vanes \cite{elston2015overview}, would allow for direct measurement of side-slip and angle of attack. While expensive and complex, these measurements would aid in aircraft modelling and control with MPC-based algorithms. Furthermore, they would also enable the MPC algorithms to fly missions in regions with strong vertical winds, such as cliffed coasts.

Second, improved thrust characterization could also be achieved via wind tunnel experiments \cite{coates2019propulsion}. While wind tunnels of the scale needed to fit a RAAVEN are sparse and expensive, accurate ground-truth characterization of the thrust curve would eliminate several variables from the grey-box system identification problem, likely simplifying and improving the lift and drag curve fits.

Finally, a common challenge with MPC algorithms is tuning the cost parameters to achieve the desired closed-loop performance. In this work, parameters were tuned offline in simulation, then ported over to the aircraft directly, with no re-tuning during flight. Manual tuning in flight is challenging and flight time is expensive. Therefore, research into automated tuning methods for MPC which run efficiently in flight are an appealing next step for this work \cite{loquercio2022autotune}.

\section{Conclusion}

 This work developed and tested two novel MPC-based guidance algorithms for high-performance path following over smooth 3D curves. To enable MPC, results from control-augmented aircraft modelling of the RAAVEN FW-sUAS directly from flight data were presented. The control-augmented model balanced predictive power with computational cost, and proved suitable for MPC guidance. The two MPC guidance algorithms, CR-MPC and MPCC, were designed for simultaneous lateral and longitudinal guidance via roll, pitch, and throttle setpoint commands to the underlying autopilot. CR-MPC regulated the aircraft to progress along the path at a fixed ground speed, while MPCC attempts to simultaneously maximize airspeed while minimizing path error, trading off these two objectives as needed. Both controllers effectively utilized soft constraints to keep the aircraft in a safe flight regime, despite noisy measurements and plant-model mismatch.

 Future work will consider additional developments to improve sensing and modelling onboard the FW-sUAS, as discussed in Section \ref{section:discussion}. We are also interested in combining MPC guidance with machine learning via meta-level policies to reduce computational overhead while improving closed-loop performance.

\section*{Acknowledgment}
\addcontentsline{toc}{section}{Acknowledgment}
The authors gratefully acknowledge hardware and flight testing support from Michael Rhodes, Jenna Cooper, Ceu Gomez-Faulk, and Maggie Wussow.

\bibliographystyle{IEEEtran}
\bibliography{refs}

\end{document}

%% file: performance_metric_table.tex
\begin{table*}[]
\centering
\caption{Performance statistics of guidance algorithms over two laps of each test path.}
\label{tab:performance_metrics}
\resizebox{\textwidth}{!}{%
\begin{tabular}{lllllllllllllllll}
\multicolumn{17}{c}{\textbf{Path-Following Error {[}m{]}}} \\ \hline
\multicolumn{1}{l|}{} &
  \multicolumn{4}{c|}{Path 1} &
  \multicolumn{4}{c|}{Path 2} &
  \multicolumn{4}{c|}{Path 3} &
  \multicolumn{4}{c}{Path 4} \\
\multicolumn{1}{l|}{Controller:} &
   &
  \multicolumn{1}{c}{Mean} &
  \multicolumn{1}{c}{Median} &
  \multicolumn{1}{c|}{Max} &
   &
  \multicolumn{1}{c}{Mean} &
  \multicolumn{1}{c}{Median} &
  \multicolumn{1}{c|}{Max} &
   &
  \multicolumn{1}{c}{Mean} &
  \multicolumn{1}{c}{Median} &
  \multicolumn{1}{c|}{Max} &
   &
  Mean &
  Median &
  Max \\ \hline
\multicolumn{1}{l|}{MPCC} &
   &
  1.077 &
  0.625 &
  \multicolumn{1}{l|}{12.906} &
   &
  6.984 &
  1.708 &
  \multicolumn{1}{l|}{43.210} &
   &
  2.553 &
  1.203 &
  \multicolumn{1}{l|}{29.071} &
   &
  2.272 &
  0.749 &
  24.958 \\
\multicolumn{1}{l|}{CR-MPC} &
   &
  1.430 &
  0.918 &
  \multicolumn{1}{l|}{12.564} &
   &
  6.372 &
  2.247 &
  \multicolumn{1}{l|}{37.599} &
   &
  2.994 &
  1.933 &
  \multicolumn{1}{l|}{17.446} &
   &
  1.964 &
  0.632 &
  15.767 \\
\multicolumn{1}{l|}{Lookahead} &
   &
  4.647 &
  3.274 &
  \multicolumn{1}{l|}{19.031} &
   &
  13.245 &
  4.337 &
  \multicolumn{1}{l|}{61.960} &
   &
  8.464 &
  4.349 &
  \multicolumn{1}{l|}{35.771} &
   &
  5.190 &
  3.021 &
  35.611 \\
 &
   &
   &
   &
   &
   &
   &
   &
   &
   &
   &
   &
   &
   &
   &
   &
   \\
\multicolumn{17}{c}{\textbf{Airspeed {[}m/s{]}}} \\ \hline
\multicolumn{1}{l|}{} &
  \multicolumn{4}{c|}{Path 1} &
  \multicolumn{4}{c|}{Path 2} &
  \multicolumn{4}{c|}{Path 3} &
  \multicolumn{4}{c}{Path 4} \\
\multicolumn{1}{l|}{Controller:} &
   &
  \multicolumn{1}{c}{Mean} &
  \multicolumn{1}{c}{Median} &
  \multicolumn{1}{c|}{Max} &
   &
  \multicolumn{1}{c}{Mean} &
  \multicolumn{1}{c}{Median} &
  \multicolumn{1}{c|}{Max} &
   &
  \multicolumn{1}{c}{Mean} &
  \multicolumn{1}{c}{Median} &
  \multicolumn{1}{c|}{Max} &
   &
  Mean &
  Median &
  Max \\ \hline
\multicolumn{1}{l|}{MPCC} &
   &
  23.542 &
  23.375 &
  \multicolumn{1}{l|}{27.886} &
   &
  23.560 &
  23.056 &
  \multicolumn{1}{l|}{31.381} &
   &
  23.815 &
  23.287 &
  \multicolumn{1}{l|}{30.284} &
   &
  26.217 &
  26.837 &
  35.671 \\
\multicolumn{1}{l|}{CR-MPC} &
   &
  25.049 &
  25.492 &
  \multicolumn{1}{l|}{29.597} &
   &
  24.231 &
  23.468 &
  \multicolumn{1}{l|}{31.820} &
   &
  25.043 &
  25.317 &
  \multicolumn{1}{l|}{29.238} &
   &
  24.386 &
  24.408 &
  29.884 \\
\multicolumn{1}{l|}{Lookahead} &
   &
  21.333 &
  21.321 &
  \multicolumn{1}{l|}{22.873} &
   &
  21.205 &
  21.202 &
  \multicolumn{1}{l|}{22.525} &
   &
  21.481 &
  21.021 &
  \multicolumn{1}{l|}{24.466} &
   &
  21.015 &
  21.016 &
  23.490 \\
 &
   &
   &
   &
   &
   &
   &
   &
   &
   &
   &
   &
   &
   &
   &
   &
   \\
\multicolumn{17}{c}{\textbf{Ground Speed {[}m/s{]}}} \\ \hline
\multicolumn{1}{l|}{} &
  \multicolumn{4}{c|}{Path 1} &
  \multicolumn{4}{c|}{Path 2} &
  \multicolumn{4}{c|}{Path 3} &
  \multicolumn{4}{c}{Path 4} \\
\multicolumn{1}{l|}{Controller:} &
   &
  \multicolumn{1}{c}{Mean} &
  \multicolumn{1}{c}{Median} &
  \multicolumn{1}{c|}{Max} &
   &
  \multicolumn{1}{c}{Mean} &
  \multicolumn{1}{c}{Median} &
  \multicolumn{1}{c|}{Max} &
   &
  \multicolumn{1}{c}{Mean} &
  \multicolumn{1}{c}{Median} &
  \multicolumn{1}{c|}{Max} &
   &
  Max &
  Median &
  Max \\ \hline
\multicolumn{1}{l|}{MPCC} &
   &
  23.351 &
  23.340 &
  \multicolumn{1}{l|}{28.186} &
   &
  23.245 &
  22.826 &
  \multicolumn{1}{l|}{33.617} &
   &
  23.286 &
  22.663 &
  \multicolumn{1}{l|}{33.521} &
   &
  25.908 &
  25.705 &
  36.300 \\
\multicolumn{1}{l|}{CR-MPC} &
   &
  24.952 &
  25.044 &
  \multicolumn{1}{l|}{26.631} &
   &
  24.059 &
  24.786 &
  \multicolumn{1}{l|}{28.888} &
   &
  24.912 &
  24.994 &
  \multicolumn{1}{l|}{29.157} &
   &
  24.182 &
  24.799 &
  26.257 \\
\multicolumn{1}{l|}{Lookahead} &
   &
  20.979 &
  20.135 &
  \multicolumn{1}{l|}{25.143} &
   &
  20.893 &
  19.764 &
  \multicolumn{1}{l|}{25.383} &
   &
  20.815 &
  19.727 &
  \multicolumn{1}{l|}{28.084} &
   &
  20.569 &
  20.259 &
  26.356 \\
 &
   &
   &
   &
   &
   &
   &
   &
   &
   &
   &
   &
   &
   &
   &
   &
   \\
\multicolumn{17}{c}{\textbf{Feedback Time {[}ms{]}}} \\ \hline
\multicolumn{1}{l|}{} &
  \multicolumn{4}{c|}{Path 1} &
  \multicolumn{4}{c|}{Path 2} &
  \multicolumn{4}{c|}{Path 3} &
  \multicolumn{4}{c}{Path 4} \\
\multicolumn{1}{l|}{Controller:} &
   &
  \multicolumn{1}{c}{Mean} &
  Median &
  \multicolumn{1}{c|}{Max} &
   &
  \multicolumn{1}{c}{Mean} &
  Median &
  \multicolumn{1}{c|}{Max} &
   &
  \multicolumn{1}{c}{Mean} &
  Median &
  \multicolumn{1}{c|}{Max} &
   &
  Mean &
  Median &
  Max \\ \hline
\multicolumn{1}{l|}{MPCC} &
   &
  \multicolumn{1}{c}{20.884} &
  20.089 &
  \multicolumn{1}{l|}{53.781} &
   &
  \multicolumn{1}{c}{23.175} &
  21.915 &
  \multicolumn{1}{l|}{43.069} &
   &
  \multicolumn{1}{c}{21.407} &
  20.757 &
  \multicolumn{1}{l|}{37.823} &
   &
  20.506 &
  19.148 &
  58.990 \\
\multicolumn{1}{l|}{CR-MPC} &
   &
  \multicolumn{1}{c}{16.121} &
  15.802 &
  \multicolumn{1}{l|}{55.494} &
   &
  \multicolumn{1}{c}{17.446} &
  16.564 &
  \multicolumn{1}{l|}{33.051} &
   &
  \multicolumn{1}{c}{15.960} &
  15.518 &
  \multicolumn{1}{l|}{36.952} &
   &
  15.688 &
  14.846 &
  34.452 \\
\multicolumn{1}{l|}{Lookahead} &
   &
  \multicolumn{1}{c}{4.935} &
  4.761 &
  \multicolumn{1}{l|}{15.199} &
   &
  \multicolumn{1}{c}{4.848} &
  4.728 &
  \multicolumn{1}{l|}{13.424} &
   &
  \multicolumn{1}{c}{4.613} &
  4.531 &
  \multicolumn{1}{l|}{9.103} &
   &
  4.644 &
  4.525 &
  11.953
\end{tabular}%
}
\end{table*}